\documentclass[12pt]{article}
\usepackage{graphicx, epsfig}
\usepackage{amsmath}
\usepackage{latexsym}
\usepackage{amstext}
\usepackage{array}
\usepackage{multirow}
\usepackage{tikz}
\usetikzlibrary{arrows}
\usepackage{changepage}
\usepackage{xfrac}
\usepackage{subcaption}
\usepackage{hyperref}

\usepackage{amsmath, amssymb}
\usepackage{pgfplots}
\pgfplotsset{compat=1.17}
\usepackage{graphicx}
\usepackage{caption}
\usepackage{subcaption}
\usepackage{float}
\usepackage{algorithm}
\usepackage{algpseudocode}
\usepackage{geometry}
\newcommand{\ds}{\displaystyle}
\newcommand{\beq}{\begin{eqnarray}}
\newcommand{\eeq}{\end{eqnarray}}
\newcommand{\beqq}{\begin{eqnarray*}}
\newcommand{\eeqq}{\end{eqnarray*}}

\newcommand{\erf}{\mbox{\boldmath$erf$}}

\font\bb=msbm10 at 12pt

\def\rR{\hbox{\bb R}}

\def\eE{\hbox{\bb E}}
\begin{document}

%%%%%%%%%%%%%%%%%%%%%%%%%%%%%%%%%%%%%%%%%%%%%%%%%%%%%%%%%%%%%%%%%%%%%%%%%%%%
\begin{center}
\Large \textbf{Assessing One‐Dimensional Cluster Stability by Extreme‐Point Trimming}\\
%\vspace{0.5cm}
%\large {\bf Section: Computational methods and algorithms}\\
\vspace{0.25cm}
%{\bf Springer Nature book about Neuromethods}\\
\vspace{0.25cm}
\normalsize
E. Dereure
\footnote{Group of Applied Mathematics and Computational Biology, Ecole Normale Sup\'erieure, PSL University, Paris, France.}, E. Akame Mfoumou$^{1}$,  and D. Holcman$^{1,}$\footnote{Churchill College, Cambridge University, CB30DS UK.}
\end{center}
%%%%%%%%%%%%%%%%%%%%%%%%%%%%%%%%%%%%%%%%%%%%%%%%%%%%%%%%%%%%%%%%%%%%%%%
\begin{abstract}
We develop a probabilistic method for assessing the tail behavior and geometric stability of one-dimensional n i.i.d. samples by tracking how their span contracts when the most extreme points are trimmed. Central to our approach is the diameter-shrinkage ratio,  that quantifies the relative reduction in data range as extreme points are successively removed. We derive analytical expressions, including finite-sample corrections, for the expected shrinkage under both the uniform and Gaussian hypotheses, and establish that these curves remain distinct even for moderate number of removal. We construct an elementary decision rule that assigns a sample to whichever theoretical shrinkage profile it most closely follows. This test achieves higher classification accuracy than the classical likelihood-ratio test in small-sample or noisy regimes, while preserving asymptotic consistency for large n. We further integrate our criterion into a clustering pipeline (e.g.\ DBSCAN), demonstrating its ability to validate one-dimensional clusters without any density estimation or parameter tuning. This work thus provides both theoretical insight and practical tools for robust distributional inference and cluster stability analysis.
\end{abstract}
%%%%%%%%%%%%%%%%%%%%%%%%%%%%%%5
\section{Introduction} \label{sec:intro}
%%%%%%%%%%%%%%%%%%%%%%%%%%5%%%%
The automated identification of clusters or isolated points is a fundamental step in many classification and spatial analysis pipelines \cite{parutto2022high,parutto2023single,perochon2025unraveling} to identify structures in unlabeled data. Clustering typically begins by assigning labels to data points, indicating their membership to one or more groups. However, the strategies used to define these groups can vary significantly across clustering methods, depending on the underlying assumptions about data structure, density, or similarity. \\
Clustering and classification algorithms can be broadly categorized into partitioning-based, hierarchical, and density-based methods. Partitioning methods, such as K-means \cite{lloyd1982least, macqueen1967some}, Spectral Clustering \cite{ng2001spectral}, and Support Vector Machines (SVMs) \cite{cortes1995support}, divide the data into distinct groups by optimizing specific criteria. K-means partitions data into a fixed number of spherical clusters by minimizing within-cluster variance. Spectral Clustering extends partitioning by leveraging the eigenstructure of similarity graphs to identify clusters with complex, non-convex shapes through an embedding step followed by a partitioning algorithm. Similarly, SVMs perform classification by implicitly mapping data into higher-dimensional feature spaces using the kernel trick, effectively partitioning data through linear separation in that transformed space. Unlike partitioning methods, hierarchical clustering builds a nested structure of clusters either from the bottom up (agglomerative, like Ward's method \cite{ward1963hierarchical} or from the top down (divisive \cite{kaufman2009finding}). In contrast, density-based methods, such as DBSCAN \cite{ester1996density}, identify clusters as dense regions of points separated by areas of lower density, allowing the discovery of arbitrarily shaped clusters and handling noise naturally, without needing to predefine the number of clusters.\\
 Together, these techniques offer complementary strengths, depending on the shape, scale, and noise level of the data. 
In applications where point clouds represent high-dimensional embeddings—such as cell representations extracted from deep learning models—unsupervised clustering plays a crucial role in uncovering latent structure prior to any downstream statistical inference. Identifying clusters automatically in such settings enables spatial statistical analyses to infer local organizational principles, as in \cite{perochon2025unraveling}, where clusters correspond to cellular subtypes or functional microenvironments used to study how subtypes colonize brain area. Identifying clustering across a large population is thus a key step to generate the statitics.\\
A persistent challenge, however, is the lack of an absolute definition of what constitutes a “cluster.” Indeed, in many spatial statistics frameworks, a cluster is meaningful only in contrast to a reference null model, often based on homogeneous Poisson or binomial point processes \cite{baddeley2015spatial}. \\
Two primary approaches have been widely employed to determine whether a given cluster represents a true hotspot or merely a spurious pattern. The frequentist approach typically relies on hypothesis testing, often implemented via Monte Carlo simulations, to assess whether the null hypothesis can be rejected \cite{xie2019significant}. In contrast, the Bayesian approach evaluates the posterior probability of the null hypothesis relative to alternative hypotheses \cite{neill2005bayesian}. For both paradigms, significant efforts have been made to reduce computational costs, leading to various optimization strategies and algorithmic improvements \cite{neill2004rapid, makatchev2008learning, xie2020unified}.\\
Recent works in computational geometry suggest alternative, model-free methods to assess the stability of point ensembles. For instance, the evolution of convex hulls \cite{barany2008random,devroye2011limit}, Voronoi cells \cite{okabe2009spatial}, or nearest-neighbor graphs \cite{penrose2003random} can provide robust geometric indicators of underlying structure. \\
We focus here on examining the stability of an ensemble under the removal of extreme points—i.e., those furthest from the center of mass. This perspective has also emerged in clustering robustness studies and outlier detection literature \cite{hennig2007clusterwise,rousseeuw2005robust}. The present manuscrit is focusing in one dimension, ans we shall focus on how a segment length (diameter) and center of mass of a point cloud evolve as extreme points are removed successively. Intuitively, point clouds drawn from long-tailed distributions (e.g., Gaussian) are expected to undergo sharper geometric changes compared to those from compact distributions (e.g., uniform). We study here  the evolution of geometric quantities for i.i.d. points in one dimension drawn from either a uniform distribution \( \mathcal{U}[0, L] \) or a Gaussian distribution \( \mathcal{N}(0, \sigma^2) \). Specifically, in \autoref{sec:order_statistics}, we derive the distribution and moments of order statistics for both the uniform and Gaussian cases. We particularly examine the expected shift in the center of mass and the behavior of the segment length (diameter) as extreme points are successively removed. In \autoref{sec:statistics_test}, we introduce a family of geometric test statistics designed to discriminate between uniform and Gaussian samples. In particular, we propose an algorithm to assess the geometric stability of empirical one-dimensional distributions, which we benchmark against the likelihood ratio test. Finally, in \autoref{sec:application_1d}, we demonstrate how this algorithm can be applied to clusters identified by a spatial clustering algorithm. \\
These statistical test can be used for cluster validation, distributional testing, or as preprocessing steps for noise filtering and robust clustering. We show analytically and numerically that Gaussian-distributed ensembles are more sensitive to the removal of extreme points, revealing an intrinsic geometric fragility compared to uniformly distributed samples. This work bridges computational geometry, statistical inference, and robust clustering, and contributes to a growing literature on model-free geometric descriptors for point cloud analysis \cite{heinrich2022extreme,brunet2007clustering,steinley2006k}.
%%%%%%%%%%%%%%%%%%%%%%%%%%%%%%%%%%%%%%%%%%%%%%%%%%%%%%%
\section{Order statistics, center of mass, and segment stability under point removal} \label{sec:order_statistics}
%%%%%%%%%%%%%%%%%%%%%%%%%%%%%%%%%%%%%%%%%%%%%5
In many statistical applications, particularly in unsupervised classification and geometric inference, a recurring question is whether an observed ensemble of points exhibits structural stability or reflects the influence of heavy-tailed variability. Motivated by this, we propose a method to assess the stability of point clouds in one dimension by iteratively removing extremal points—those furthest from the ensemble’s center of mass—and tracking the resulting geometric changes.\\
We start with \( \mathcal{S}_n = \{x_1, \dots, x_n\} \subset \mathbb{R} \) which is a sample of \( n \) i.i.d. real-valued random variables. To formulate a statistical criterion for assessing the underlying distribution of such a sample, we introduce a procedure called the \textit{Diameter Shrinkage Statistic}, which quantifies how the total length (i.e., the diameter) of the ensemble evolves as we remove the most extreme values.\\
We begin by defining the ordered sample \( \mathcal{O}_n = \{X_{(1)} \leq \dots \leq X_{(n)}\} \), and for any integer \( 0 \leq p < n/2 \), we consider the segment length:
\[
D_p := X_{(n-p)} - X_{(p+1)},
\]
corresponding to the length of the central segment obtained after removing the \( p \) smallest and \( p \) largest points from the sample. We define the diameter shrinkage ratio at step \( p \) as:
\beq \label{eq:def_shrinkage_step_p}
    T_{\text{shrink}}^{(p)} := \frac{D_p}{D_{p-1}},
\eeq
which reflects the relative contraction of the ensemble’s diameter due to the exclusion of the next outermost pair of points. Our goal is to characterize the distribution and expected value of this statistic under two reference models: the uniform distribution \( \mathcal{U}[a, b] \) and the normal distribution \( \mathcal{N}(\mu, \sigma^2) \).\\
In the scope of this work, to initiate the study of the statistics $T_{\text{shrink}}^{(p)}$, rather than computing the exact expected value, we approximate the expected shrinkage ratio via the ratio of expected lengths (somehow justified by \eqref{eq:esperance_evolution_ratio_segment_length_xn_justify_approx} in Appendix-A:
\beq \label{eq:expectation_approx}
    \mathbb{E}(T_{\text{shrink}}^{(p)}) \approx \frac{\mathbb{E}(D_p)}{\mathbb{E}(D_{p-1})},
\eeq
thereby reducing the analysis to that of the order statistics \( X_{(k)} \) for \( 1 \leq k \leq n \). This approximation can be further examined in future work. In the next subsection, we compute formula \ref{eq:expectation_approx} in the case of a Uniform Distribution statistics.
%%%%%%%%%%%%%%%%%%%%%%%%%%%%%%%%%%%%%%5
\subsection{Shrinkage length ratio for a uniform Distribution}
%%%%%%%%%%%%%%%%%%%%%%%%%%%%%%%%%%%%%%%5
We first consider the case where the sample is drawn from the uniform distribution on the interval \( [0, L] \). Rigorous computations extreme statistics are available \cite{karlin1981second, david2004order, majumdar2024statistics} and we recall here elementary derivation to make this manuscript self contained. We recall that the joint probability density function of the ordered sample \( X_{(1)}, \dots, X_{(n)} \) is given by:
\begin{equation} \label{eq:density_uniform}
  f(x_1, \dots, x_n) =
  \begin{cases}
    \dfrac{n!}{L^n} & \text{if } 0 \leq x_1 \leq \dots \leq x_n \leq L, \\
    0 & \text{otherwise}.
  \end{cases}
\end{equation}
The marginal density of the \( k \)-th order statistic \( X_{(k)} \) can be computed by integrating this density over all the other variables under the ordering constraints:
\beq
f_k(x_k) & =\int_0^L \dots  \int_0^L f(x_1, \dots, x_k, \dots, x_n) dx_1 \dots dx_{k-1} dx_{k+1} \dots dx_n \\
& = \frac{n!}{L^n} \left(\int_0^{x_k} dx_1 \int_{x_1}^{x_k}dx_2 \dots  \int_{x_{k-2}}^{x_k}dx_{k-1}\right) \left(\int_{x_k}^L d_{x_{k+1}} \int_{x_{k+1}}^L d_{x_{k+2}} \dots \int_{x_{n-1}}^L dx_{n}\right) \\
&= \frac{n!}{L^n} L(x_k)  R(x_k),
\eeq 
where the terms \( L(x_k) \) and \( R(x_k) \) respectively encode the cumulative integration over the \( k-1 \) variables smaller than \( x_k \), and the \( n-k \) variables greater than \( x_k \). These terms can be computed by successive integrations of a polynomial function:
\beq 
& L(x_k) = \int_0^{x_k} dx_1 \int_{x_1}^{x_k}dx_2 \dots  \int_{x_{k-2}}^{x_k}dx_{k-1} = \frac{x_k^{k-1}}{(k-1)!}\\
& R(x_k) = \int_{x_k}^L d_{x_{k+1}} \int_{x_{k+1}}^L d_{x_{k+2}} \dots \int_{x_{n-1}}^L dx_{n} = \frac{(L-x_k)^{n-k}}{(n-k)!}.
\eeq 
The marginal density of the  \( k \)-th order statistic \( X_{(k)} \) is given by:
\begin{equation} \label{eq:distribution_xk}
    f_k(x) = \frac{n!}{(k-1)!(n-k)!} \cdot \frac{x^{k-1}(L - x)^{n - k}}{L^n}, \quad x \in [0, L].
\end{equation}
The density function in Equation~\eqref{eq:distribution_xk} is a scaled Beta distribution on interval $[0, L]$. Since the Beta $\ds B$ and Gamma functions are connected by $\ds B(\alpha, \beta) = \frac{\Gamma(\alpha)\Gamma(\beta)}{\Gamma(\alpha+\beta)}$, the expected value of the \( X_{(k)} \) variable is given by:
\beq \label{eq:expectation_xk}
\mathbb{E}(X_{(k)}) = \frac{k}{n+1} \cdot L
\eeq 
The details of the computations are presented in Appendix-A see also \cite{karlin1981second, david2004order, majumdar2024statistics}. Using this expression, the expected segment length shortening after removal of the \( p \) smallest and \( p \) largest points becomes:
\[
\mathbb{E}(D_p) = \mathbb{E}(X_{(n - p)}) - \mathbb{E}(X_{(p+1)}) = \frac{n - 2p - 1}{n+1} \cdot L.
\]
Consequently, the expected shrinkage ratio satisfies:
\begin{equation} \label{eq:shrinkage_ratio_uniform}
    \mathbb{E}(T_{\text{shrink}}^{(p)}) \approx \frac{n - 2p - 1}{n - 2p + 1}.
\end{equation}
Expression \ref{eq:shrinkage_ratio_uniform}  is valid regardless of the bounds of the uniform distribution $\mathcal{U}[a, b]$. This provides a baseline expectation under the uniform model. Deviations from this behavior-—particularly under heavy-tailed distributions such as the Gaussian—-can then be used to construct hypothesis testing or stability metrics. In the following sections, we extend this analysis to Gaussian samples and demonstrate how the shrinkage statistic behaves differently, thus enabling the design of robust geometric tests for distributional classification.
%%%%%%%%%%%%%%%%%%%%%%%%%%%%%%%%%%%%%%5
\subsection{Shift of the Sample Mean when Removing an Extreme Uniform Outlier}
%%%%%%%%%%%%%%%%%%%%%%%%%%%%%%%%%%%%%%%%%%%%
We now quantify how the sample mean (center of mass) is shifted when we discard the single largest observation from \(n\) i.i.d.\ draws \(X_{(1)}\le\cdots\le X_{(n)}\sim\mathcal U[0, L]\). By definition,
\[
  \bar X
  = \frac1n\sum_{i=1}^n X_{(i)},
  \quad
  \mathbb{E}[\bar X]
  = \frac1n\sum_{i=1}^n\mathbb{E}[X_{(i)}]
  = \frac{n(n+1)L}{2n(n+1)} = \frac{L}{2}
  \]
%%%%%%%%%%%%%%%%%%%%%%%%%%%%%%%%%%%
\paragraph{Trimmed mean after removing the maximum.}
Delete \(X_{(n)}\) and form the \((n-1)\)-point mean
\(\displaystyle
  \bar X_{-}
  = \frac1{n-1}\sum_{i=1}^{n-1}X_{(i)}.
\)
Its expectation is
\[
  \mathbb{E}[\bar X_{-}]
  = \frac1{n-1}\sum_{i=1}^{n-1}\mathbb{E}[X_{(i)}] = \frac{n(n-1)}{2(n-1)(n+1)}=\frac{nL}{2(n+1)}.
\]

Therefore
\[
  \mathbb{E}[\bar X - \bar X_{-}] = \frac{L}{2} - \frac{nL}{2(n+1)} = \frac{L}{2(n+1)}.
\]
%%%%%%%%%%%%%%%%%%%%%%%%%%%%%%%%%%%
\paragraph{Net shift.}
The expected shift in the center of mass upon removal of the single largest point is thus
\[
  \mathbb{E}\bigl[\bar X - \bar X_{-}\bigr]
  = \frac{L}{2(n+1)}.
\]

This expression stays valid regardless of the bounds of the uniform distribution $\mathcal{U}[a,b]$ by switching $L$ by $b-a$. By an identical argument for the smallest point—and by symmetry—the simultaneous removal of both extremes produces no first‐order shift in the mean.

%%%%%%%%%%%%%%%%%%%%%%%
\subsection{Shrinkage length ratio for Gaussian Distribution}
%%%%%%%%%%%%%%%%%%%55
We now repeat the approach developed above to the case of a sampling \( \mathcal{S}_n = \{x_1, \dots, x_n\} \subset \mathbb{R} \)  drawn for i.i.d. from a normal distribution \( \mathcal{N}(0, \sigma^2) \). In contrast to the uniform case, the analysis of order statistics for Gaussian variables is considerably more involved, as the literature providing closed-forme expressions for such order statistics distributions \cite{biroli2024exact} often relies on general and technically sophisticated theorems like the Fisher-Tippett-Gnedenko theorem. Here, we aim to recover similar results using more elementary and self-contained computations, in the hope of providing a more accessible perspective. By combining integral decompositions with properties of the Gaussian cumulative and density functions, we will see below that we can derive close expressions.
%%%%%%%%%%%%%%%%%%%%%%%%%%55
\subsubsection{Approximated Distribution of $|S^{max}_n|$} \label{sec:distribution_max_abs_gaussian}
%%%%%%%%%%%%%%%%%%%%%%%%%%%5
We present here an elementary asymptotic expression for the mean distance of the extreme point among n i.i.d Gaussian variable in $\rR$. Thus \(S_1,\dots,S_n\) be i.i.d.\ \(\mathcal N(0,\sigma^2)\), and denote
\beq
|S_n^{\max}| \;=\;\max_{1\le k\le n} |S_k|\,. 
\eeq
Our aim is to  show from elementary computations that 
\beq \label{eq:expected_value_absolute_gaussian}
\mathbb{E}\bigl[\,|S_n^{\max}|\bigr] \;\sim\ \sigma\sqrt{\tfrac{\pi\,\ln n}{2}}.
\eeq
We start with the identify for \(R\ge0\),
\[
  \Pr\{|S|>R\}
  =1-\Pr\{|S|\le R\}
  =1-\erf\!\Bigl(\tfrac{R}{\sigma\sqrt2}\Bigr),
\]
where 
\(\displaystyle\erf(x)=\frac{2}{\sqrt\pi}\int_0^x e^{-t^2}dt\). Since \(\{|S_k|\le R\}\) are independent events,
\[
  \Pr\{|S_n^{\max}|>R\}
  =1-\Pr\{|S|\le R\}^n
  =1-\erf\!\Bigl(\tfrac{R}{\sigma\sqrt2}\Bigr)^n.
\]
In addition, 
\[
  \mathbb{E}\bigl[|S_n^{\max}|\bigr]
  =\int_0^\infty \Pr\{|S_n^{\max}|>r\}\,dr
  =\int_0^\infty
    \Bigl[\,1-\erf\!\bigl(\tfrac{r}{\sigma\sqrt2}\bigr)^n\Bigr]
  \,dr.
\]
Using the approximation of \(\erf\) \cite{chu1955distribution} (see Appendix-B for different asymptotic computation)
$  \erf(x)\approx\sqrt{1-e^{-4x^2/\pi}}, x\ge0.$
Thus
\[
  \erf\!\bigl(\tfrac{r}{\sigma\sqrt2}\bigr)^n
  \approx
  \Bigl(1-e^{-\,\tfrac{2r^2}{\pi\sigma^2}}\Bigr)^{\!n/2}.
\]
Set
\[
  I_n
  =\int_0^\infty
    \Bigl[\,1-\bigl(1-e^{-2r^2/(\pi\sigma^2)}\bigr)^{n/2}\Bigr]
  \,dr,
\]
so \(\mathbb{E}[|S_n^{\max}|]\approx I_n\).
We now compute the integral by a first change of variables.
\[
  u=\sqrt{\tfrac{2}{\pi\sigma^2}}\;r,
  \quad
  dr=\sigma\,\sqrt{\tfrac\pi2}\;du.
\]
Then
\[
  I_n
  =\sigma\sqrt{\tfrac\pi2}
   \int_0^\infty
     \Bigl[\,1-(1-e^{-u^2})^{n/2}\Bigr]
   \,du.
\]
followed by a second change of variables \(u=v\sqrt{\ln n}\), so \(du=\sqrt{\ln n}\,dv\).  Hence
\[
  I_n
  =\sigma\sqrt{\tfrac{\pi\,\ln n}{2}}
   \int_0^\infty
     \Bigl[\,1-\bigl(1-n^{-v^2}\bigr)^{n/2}\Bigr]
   \,dv.
\]
Finally, we obtain the asymptotic via dominated convergence: for each fixed \(v\),
\[
  \bigl(1-n^{-v^2}\bigr)^{n/2}
  =\exp\bigl(\tfrac n2\ln(1-n^{-v^2})\bigr)
  \;\longrightarrow\;
  \begin{cases}
    0,&0\le v<1,\\
    1,&v>1,
  \end{cases}
\]
as \(n\to\infty\).  Since the integrand is bounded by 1, dominated convergence yields
\[
  \int_0^\infty
    \Bigl[1-(1-n^{-v^2})^{n/2}\Bigr]\,dv
  \;\longrightarrow\; 1.
\]
Therefore
\[ 
  \mathbb{E}[|S_n^{\max}|]\sim I_n\sim\sigma\sqrt{\tfrac{\pi\,\ln n}{2}}.
\]
Although, in the literature, the large n limit gives 
\beq \label{eq:literature_expected_value}
  \mathbb{E}[|S_n^{\max}|] \sim \mathbb{E}[X_n] \sim \sigma\sqrt{2\log n},
\eeq 
 \cite{biroli2024exact} and Appendix-B. However, \eqref{eq:literature_expected_value} is the asymptotic expression, for relatively small values of $n$ ($n \leq 10,000$), and the present formula \eqref{eq:expected_value_absolute_gaussian} is closest to the simulations for n not that large. We therefore will use \eqref{eq:expected_value_absolute_gaussian} in the following of this work.
%%%%%%%%%%%%%%%%%%%%%%%%%%%%%%%%555
\begin{figure*}[!h]
\centerline{\includegraphics[width=1\columnwidth]{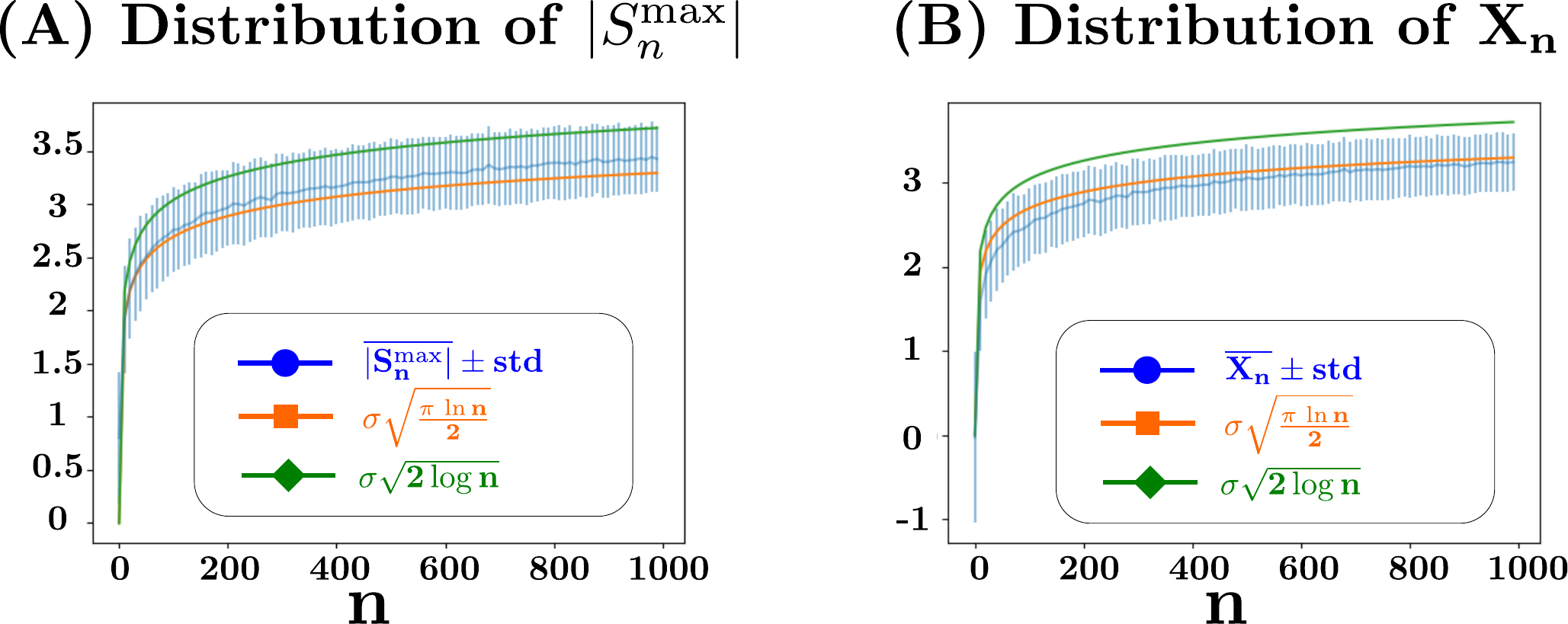}}
\caption{Distribution of the maximum among $n$ i.i.d. Gaussian variable in $\mathbb{R}$. (A): Distribution of the distance of the extreme point. (B): Distribution of the maximum. We used $\sigma=1$ and for each $n$ the simulations are averaged over $1,000$ runs to compute the average and standard deviation.}
\label{fig:statistics_maximum_gaussian}
\end{figure*}
%%%%%%%%%%%%%%%%%%%%%%%%%%%%%%%%%%%%%%%%%%%%5
%%%%%%%%%%%%%%%%%%%%%%%%%%%%%%%%%%%%%%5
%%%%%%%%%%%%%%%%%%%%%%%%%%%%%%%%%%%%%%5
%%%%%%%%%%%%%%%%%%%%%%%%%%%%%%%%%%%%%%5
\subsubsection{Probability density function function of the \( k \)-th order statistic}
%%%%%%%%%%%%%%%%%%%%%%%%%%%%%%%%%%%%%%5
We shall here derive an analytic expression for the
probability density function function of the \( k \)-th order statistic. We start with the joint density of the ordered statistics \( X_{(1)} \leq \dots \leq X_{(n)} \) for the i.i,d Gaussian variables. It is given by:
\begin{equation} \label{eq:density_gaussian_joint}
  f(x_1, \dots, x_n) =
  \begin{cases}
       \dfrac{n!}{(\sigma\sqrt{2 \pi})^n} \exp\left(-\dfrac{1}{2\sigma^2} \sum_{i=1}^n x_i^2\right) & \text{if } x_1 \leq \dots \leq x_n, \\
       0 & \text{otherwise}.
  \end{cases}
\end{equation}
The marginal density \( f_k(x_k) \) of the \( k \)-th order statistic, is obtained by integrating out all other variables under the ordering constraints:
\beq
f_k(x_k) & = \int_{-\infty}^{+\infty} \dots  \int_{-\infty}^{+\infty} f(x_1, \dots, x_k, \dots, x_n) dx_1 \dots dx_{k-1} dx_{k+1} \dots dx_n \\
& = \frac{n! e^{-\frac{x_k^2}{2\sigma^2}}}{(\sigma\sqrt{2 \pi})^n} \left(\int_{-\infty}^{x_k} \int_{-\infty}^{x_{k-1}} \dots  \int_{-\infty}^{x_2} e^{-\frac{\sum_{i=1}^{k-1} x_i^2}{2\sigma^2}} dx_1 \dots dx_{k-1}\right) \left(\int_{x_k}^{+\infty} \int_{x_{k+1}}^{+\infty}  \dots \int_{x_{n-1}}^{+\infty} e^{-\frac{\sum_{i=k}^{n} x_i^2}{2\sigma^2}} dx_{k+1} \dots dx_{n}\right) \\
&= \frac{n!e^{-\frac{x_k^2}{2\sigma^2}}}{(\sigma\sqrt{2 \pi})^n} L(x_k)  R(x_k),
\eeq
where
\beq 
L(x_k) = \int_{-\infty}^{x_k} \int_{-\infty}^{x_{k-1}} \dots  \int_{-\infty}^{x_2} e^{-\frac{\sum_{i=1}^{k-1} x_i^2}{2\sigma^2}} dx_1 \dots dx_{k-1}\\
R(x_k) = \int_{x_k}^{+\infty} \int_{x_{k+1}}^{+\infty}  \dots \int_{x_{n-1}}^{+\infty} e^{-\frac{\sum_{i=k}^{n} x_i^2}{2\sigma^2}} dx_{k+1} \dots dx_{n}.
\eeq 
Both \( L(x_k) \) and \( R(x_k) \) respectively accounts for the cumulative integration over the \( k-1 \) variables smaller than \( x_k \), and the \( n-k \) variables greater than \( x_k \). We compute now  $R(x_k)$ by induction. Indeed, 
\beq
\int_{x_{n-1}}^{+\infty} e^{-\frac{x_n^2}{2\sigma^2}}dx_n = \frac{\sigma \sqrt{2\pi}\text{erfc}(\frac{x_{n-1}}{\sigma\sqrt{2}})}{2}
\eeq
and
\beq
\int_{x_{n-2}}^{+\infty} \frac{\sigma \sqrt{2\pi}\text{erfc}(\frac{x_{n-1}}{\sigma\sqrt{2}})}{2} e^{-\frac{x_{n-1}^2}{2\sigma^2}}dx_{n-1} = \frac{(\sigma \sqrt{2\pi})^2\text{erfc}(\frac{x_{n-2}}{\sigma\sqrt{2}})^2}{8}.
\eeq
Thus, through recursive integration, we obtain the general expression 
\beq  \label{Rxk}
R(x_k) = \int_{x_k}^{+\infty} \int_{x_{k+1}}^{+\infty}  \dots \int_{x_{n-1}}^{+\infty} e^{-\frac{\sum_{i=k}^{n} x_i^2}{2\sigma^2}} dx_{k+1} \dots dx_{n} = \frac{(\sigma \sqrt{2\pi})^{n-k}\text{erfc}(\frac{x_{k}}{\sigma\sqrt{2}})^{n-k}}{2^{n-k}(n-k)!}.
\eeq 
The term $L(x_k)$ can be computed similarly using the same recursive integration: 
\beq \label{Lxk}
L(x_k) = \frac{(\sigma \sqrt{2\pi})^{k-1}\text{erfc}(-\frac{x_{k}}{\sigma\sqrt{2}})^{k-1}}{2^{k-1}(k-1)!}.
\eeq  
Finally, by combining these explicit formula \ref{Lxk}-\ref{Rxk}, we obtain an explicit expression for the marginal density:
\begin{equation} \label{eq:distribution_xk_gaussian_final}
    f_k(x_k) = \frac{n!}{(k-1)!(n-k)!} \cdot \frac{\text{erfc}\left(-\frac{x_k}{\sigma\sqrt{2}}\right)^{k-1} \text{erfc}\left(\frac{x_k}{\sigma\sqrt{2}}\right)^{n-k} \cdot e^{-\frac{x_k^2}{2\sigma^2}}}{2^{n-1} \sigma \sqrt{2\pi}}.
\end{equation}
%%%%%%%%%%%%%%%%%%%%%%%
\subsubsection{Explicit expression for  k-th order statistic ${E}[X_{(k)}]$}
%%%%%%%%%%%%%%%%%%%%%%%%%%%%%%%%%%%%%%%%%%%
To compute the expectation of the \(k\)th order statistic \(X_{(k)}\) of \(n\) i.i.d.\ \(\mathcal N(0,\sigma^2)\) samples, we start from the general identity
\[
  \mathbb{E}[X_{(k)}]
  \;=\;
  \int_{0}^{\infty}\bigl[1 - F_k(x)\bigr]\,dx
  \;-\;
  \int_{-\infty}^{0}F_k(x)\,dx,
\]
where \(F_k(x)=\int_{-\infty}^x f_k(y)\,dy\) is the CDF of \(X_{(k)}\). By expanding the joint Gaussian density and applying the binomial, we have 
\[
  f_k(x)
  \;=\;
  \frac{n!}{(k-1)!\,(n-k)!}
  \sum_{i=0}^{n-k}
    (-1)^{\,n-k-i}
    \binom{n-k}{i}
    F(x)^{\,n-i-1}\,f(x),
\]
with \(f(x)\), \(F(x)\) the standard normal PDF and CDF.  Integrating term‐by‐term yields
\begin{align*}
  1 - F_k(x)
  &=\;
  \frac{n!}{(k-1)!\,(n-k)!}
  \sum_{i=0}^{n-k}
    (-1)^{\,n-k-i}\,
    \binom{n-k}{i}\,
    \frac{1 - F(x)^{\,n-i}}{n-i},
  \\
  F_k(x)
  &=\;
  \frac{n!}{(k-1)!\,(n-k)!}
  \sum_{i=0}^{n-k}
    (-1)^{\,n-k-i}\,
    \binom{n-k}{i}\,
    \frac{F(x)^{\,n-i}}{n-i}.
\end{align*}
In the regime \(n-k\ll n\), the lower‐tail integral \(\int_{-\infty}^0F_k(x)\,dx\) is negligible, and for \(x\ge0\), \(F(x)^{\,n-i}\) is exponentially small.   We shall approximate the expectation by (see Appendix \ref{sec:AppendixGaussian}):
\begin{equation}
\mathbb{E}(X_{(k)}) \approx \int_0^{+\infty} (1 - F_k(x)) dx \approx \frac{n!}{(k-1)!(n-k)!} \sum_{i=0}^{n-k} \binom{n-k}{i} \frac{(-1)^{n-k-i}}{n-i} \cdot \frac{\sigma\sqrt{2\pi \ln(n-i)}}{2}.
\end{equation}
We conclude that for \( k \approx n \), the mean position of the k$^{th}$ order position of n Gaussian i.i.d variable is located at:
\begin{equation} \label{eq:expected_value_xk_gaussianp}
    \mathbb{E}(X_{(k)}) \approx \frac{\sigma\sqrt{2\pi} \cdot n(n-1)\dots(n-(n-k))}{2} \sum_{i=0}^{n-k} \frac{(-1)^{n-k-i} \sqrt{\ln(n-i)}}{i!(n-k-i)!(n-i)}.
\end{equation}
In particular, for \(k=n-m\) with \(m\ll n\), this simplifies to
\begin{equation}\label{eq:approx_X_{(n)}_minus_m}
  \mathbb{E}[X_{(n-m)}]
  \;\approx\;
  \frac{\sigma\sqrt{2\pi}}{2}
  \;n(n-1)\cdots(n-m)
  \sum_{i=0}^m
    \frac{(-1)^{\,m-i}\,\sqrt{\ln(n-i)}}{\,i!\,(m-i)!\,(n-i)}.
\end{equation}
This expression captures both the leading $\sigma\sqrt{2\ln n}$ term and finite‐$n$ corrections via the alternating sum.  We shall now continue with further approximation of $\sqrt{\ln(n-i)}$: in the regime \(i\ll n\), we write
\[
  \sqrt{\ln(n-i)}
  = \sqrt{\ln\bigl(n(1-\tfrac{i}{n})\bigr)}
  = \sqrt{\ln n \;+\;\ln(1-\tfrac{i}{n})}
  = \sqrt{\ln n}\,\sqrt{1 \;+\;\tfrac{\ln(1-\tfrac{i}{n})}{\ln n}}.
\]
Since \(\tfrac{i}{n}\) is small, \(\ln(1-\tfrac{i}{n})=O(\tfrac{i}{n})\ll1\).  Hence we Taylor‐expand \(\sqrt{1+x}\) at \(x=0\):
\[
  \sqrt{1+x}
  = 1 + \tfrac{x}{2} + O(x^2).
\]
Taking \(x=\tfrac{\ln(1-\tfrac{i}{n})}{\ln n}\) gives the one‐term approximation
\[
  \sqrt{\ln(n-i)}
  \approx \sqrt{\ln n}
    \Bigl(1
      + \tfrac{1}{2}\,\tfrac{\ln(1-\tfrac{i}{n})}{\ln n}
    \Bigr).
\]
A further first‐order Taylor of \(\ln(1-x)\approx -x\) then yields
\[
  \sqrt{\ln(n-i)}
  \approx \sqrt{\ln n}
    \Bigl(1
      - \tfrac{1}{2}\,\tfrac{i}{n\,\ln n}
    \Bigr).
\]
Substituting into the alternating sum in 
\eqref{eq:expected_value_xk_gaussian}, where \(K=n-k\), we obtain
\[
  \mathbb{E}[X_{(k)}]
  \;\approx\;
  \sqrt{\ln n}\,\frac{\sigma\sqrt{2\pi}\,\bigl(n(n-1)\cdots(n-K)\bigr)}{2}
  \sum_{i=0}^{K}
    \frac{(-1)^{\,K-i}}{i!\,(K-i)!\,(n-i)}
    \Bigl(1 - \tfrac{i}{2n\,\ln n}\Bigr).
\]
We now simplify the two key alternating sums.  First, observe the partial‐fraction identity
\[
  \sum_{i=0}^K
    \frac{(-1)^{\,K-i}}{i!\,(K-i)!\,(n-i)}
  = \frac{1}{n(n-1)\cdots(n-K)},
\]
which follows by evaluating the decomposition
\( \ds \tfrac1{z(z-1)\cdots(z-K)}=\sum_i\frac{a_i}{z-i}\) at \(z=n\) leading to
\beq
a_j = \frac{(-1)^{K-j}}{j!(K-j)!}.
\eeq 
Second, for the sum weighted by \(i\), note
\[
  \sum_{i=0}^K
    \frac{(-1)^{\,K-i}\,i}{i!\,(K-i)!\,(n-i)}
  = \sum_{i=1}^K
    \frac{(-1)^{\,K-i}}{(i-1)!\,(K-i)!\,(n-i)}
  = \frac{1}{(n-1)(n-2)\cdots(n-K)}.
\]
Combining these two identities yields the compact approximation
\[
  \mathbb{E}[X_{(k)}]
  \;\approx\;
  \sqrt{\ln n}\,\frac{\sigma\sqrt{2\pi}}2
  \,\Biggl[
    \frac{n(n-1)\cdots(n-K)}{n(n-1)\cdots(n-K)}
    \;-\;
    \frac{n(n-1)\cdots(n-K)}{2n\,\ln n}\,\frac{1}{(n-1)\cdots(n-K)}
  \Biggr],
\]
which simplifies to
\[
    \mathbb{E}[X_{(k)}]
    \;\approx\;
    \sigma\sqrt{\tfrac{\pi\,\ln n}{2}}
    \Bigl(1 - \tfrac1{2\ln n}\Bigr).
\]
Thus to first order one recovers the familiar \(\sigma\sqrt{2\ln n}\) scaling, with a \(O\bigl((\ln n)^{-1}\bigr)\) correction. We obtain here an analytical expression that does not depend on $k$. To clarify this result, we shall now consider the second-order expansion analysis.
%%%%%%%%%%%%%%%%%%%
\subsubsection*{Higher‐order Taylor corrections}
%%%%%%%%%%%%%%%%%%%%%%%%%%%%%%%%%%%%%%%%%%%%%%5
To refine the first‐order approximation 
\(\sqrt{\ln(n-i)}\approx\sqrt{\ln n}\bigl(1-\tfrac{i}{2n\ln n}\bigr)\), 
we carry out a second‐order expansion of \(\ln(1-x)\) about \(x=0\).  Recall
\[
  \ln(1-x)
  = -\,x \;-\;\tfrac{x^2}{2}\;+\;O(x^3),
  \quad
  x=\frac{i}{n}\ll1.
\]
Hence
\[
  \ln\bigl(n-i\bigr)
  = \ln n + \ln\bigl(1-\tfrac{i}{n}\bigr)
  = \ln n
    -\frac{i}{n}
    -\frac{i^2}{2n^2}
    +O\!\bigl(n^{-3}\bigr).
\]
Taking square‐roots gives
\[
  \sqrt{\ln(n-i)}
  = \sqrt{\ln n}\,
    \sqrt{1 - \frac{i}{n\ln n} - \frac{i^2}{2n^2\ln n} + O(n^{-3})}.
\]
Expanding \(\sqrt{1+u}=1+\tfrac u2-\tfrac{u^2}{8}+O(u^3)\) with
\(u=-\tfrac{i}{n\ln n}-\tfrac{i^2}{2n^2\ln n}\) yields
\[
  \sqrt{\ln(n-i)}
  \approx
  \sqrt{\ln n}\,\Bigl[
    1
    -\tfrac{1}{2}\Bigl(\tfrac{i}{n\ln n}+\tfrac{i^2}{2n^2\ln n}\Bigr)
    -\tfrac{1}{8}\Bigl(\tfrac{i}{n\ln n}\Bigr)^2
  \Bigr].
\]
Retaining only terms up to \(O(n^{-2}\ln n ^{-1})\) gives
\beq
  {
  \sqrt{\ln(n-i)}
  \approx
  \sqrt{\ln n}\,\Bigl(
    1
    -\frac{i}{2n\ln n}
    -\frac{i^2}{4n^2\ln n}
  \Bigr).
  }
\eeq

\medskip
\noindent
Substituting this into the alternating‐sum formula
\[
  \mathbb{E}[X_{(k)}]
  \approx
  \sqrt{\ln n}\,\frac{\sigma\sqrt{2\pi}}{2}
  \sum_{i=0}^{K}
    \frac{(-1)^{K-i}}{i!\,(K-i)!\,(n-i)}
    \sqrt{\ln(n-i)},
  \quad
  K=n-k,
\]
we must evaluate the sums:
\[
  S_0
  = \sum_{i=0}^{K}\frac{(-1)^{K-i}}{i!\,(K-i)!\,(n-i)},
  \quad
  S_1
  = \sum_{i=0}^{K}\frac{(-1)^{K-i}\,i}{i!\,(K-i)!\,(n-i)},
  \quad
  S_2
  = \sum_{i=0}^{K}\frac{(-1)^{K-i}\,i^2}{i!\,(K-i)!\,(n-i)}.
\]
The standard partial‐fraction argument shows
\[
  S_0 = \frac{1}{n(n-1)\cdots(n-K)},
  \qquad
  S_1 = \frac{1}{(n-1)(n-2)\cdots(n-K)},
\]
while a similar index shift gives
\[
  S_2
  = \sum_{i=1}^K\frac{(-1)^{K-i}\,i}{(i-1)!\,(K-i)!\,(n-i)}
  = \frac{n}{(n-1)(n-2)\cdots(n-K)}.
\]
Hence the expanded expectation becomes
\[
  \mathbb{E}[X_{(k)}]
  \approx
  \frac{\sigma\sqrt{2\pi\,\ln n}}{2}
  \Bigl[
    S_0 
    -\frac{1}{2n\ln n}\,S_1
    -\frac{1}{4n^2\ln n}\,S_2
  \Bigr].
\]
Substituting the closed‐form \(S_0,S_1,S_2\) and simplifying yields
\[
  \mathbb{E}[X_{(k)}]
  \approx
  \sigma\sqrt{\tfrac{\pi\,\ln n}{2}}
  \Bigl(
    1
    -\frac{1}{2\ln n}
    -\frac{1}{4\ln n}
  \Bigr),
\]
which refines the leading \(\sigma\sqrt{2\ln n}\) term by \(O((\ln n)^{-1})\) correction. To conclude the mean position of the $k^{th}$ order position of $n$ Gaussian i.i.d. can be approximated by 
\begin{equation}
    \mathbb{E}(X_{(k)}) \approx \sqrt{\ln{n}} \frac{\sigma\sqrt{2\pi}}{2} \left(1 - \frac{1}{2\ln{(n)}}\left( 1 + \frac{1}{2}\right)\right)
\end{equation}
We obtain an approximation for \(\mathbb{E}[X_{(k)}]\) that depends only on \(n\) (not on \(k\)) once \(k\) is in the extreme tail.  In fact, carrying the Taylor expansion of \(\ln(1-x)\) to \(j\)th order shows that
\beq
  \mathbb{E}[X_{(k)}]
  \;&\approx&\;
  \frac{\sigma\sqrt{2\pi\,\ln n}}{2}
  \Bigl(1 \;-\;\frac{H_j}{2\,\ln n}\Bigr),
  \quad \\
  H_j &=& \sum_{i=1}^j \frac1i.
\eeq
To evalute the quality of this approximation, we compared, for \(n=100\) and \(\sigma=1\), the two sequences
\[
  u_k = \frac{\sqrt{2\pi\,\ln n}}{2}\Bigl(1 - \tfrac{H_{\,n-k}}{2\ln n}\Bigr),
  \quad
  v_k = \frac{\sqrt{2\pi}\,n(n-1)\cdots(n-(n-k))}{2}
         \sum_{i=0}^{n-k}
         \frac{(-1)^{\,n-k-i}\,\sqrt{\ln(n-i)}}
              {\,i!\,(n-k-i)!\,(n-i)\,},
\]
for \(k=n,n-1,\dots,n-4\).  As shown in Table~\ref{tab:comparison_approximations_e_xk}, \(\lvert u_k-v_k\rvert<4\times10^{-2}\) in all cases, demonstrating that the harmonic‐sum correction yields excellent accuracy for small \(j\).
%%%%%%%%%%%%%%%%%%%%%%%%%%%%%%%%%%%%%%%%%%%%%%5
\begin{table}[h!]
\centering
\begin{tabular}{|c|c|c|c|c|c|} 
\hline
$k$ & $$n & $n-1$ & $n-2$ & $n-3$ & $n-4$ \\
\hline
$u_k$ & $3.12389802$ & $2.87248195$ & $2.74677391$ & $2.66296856$ & $2.60011454$ \\
$v_k$ & $3.12389802$ & $2.87220937$ & $2.73608587$ & $2.64103412$ & $2.56728926$ \\
\hline
\end{tabular}
\caption{Comparison of $2$ different versions of the approximation of the expected value of the position of the $k^{th}$ order position of $n$ Gaussian i.i.d., for different values of $k$ close to $n$.} \label{tab:comparison_approximations_e_xk}
\end{table}
%%%%%%%%%%%%%%%%%%%%%%%%%%%%%%555
\subsection{Practical approximation}
%%%%%%%%%%%%%%%%%%%%%%%%%%%%%%%%%%%%%%%%55
Based on the above derivation, we propose the following practical approximation for the expected location of the \(k\)th order statistic in an i.i.d.\ sample of size \(n\) from \(\mathcal{N}(0,\sigma^2)\):
\begin{equation}\label{eq:approx_Xk_gauss}
  \mathbb{E}[X_{(k)}]
  \;\approx\;
  \frac{\sigma\sqrt{2\pi\,\ln n}}{2}
  \Bigl(1 \;-\;\frac{H_{\,n-k}}{2\,\ln n}\Bigr),
  \quad
  H_{m}=\sum_{i=1}^{m}\frac1i.
\end{equation}
Here \(H_{n-k}\) is the \((n-k)\)th harmonic number, introducing a mild finite-\(n\) correction to the leading \(\sigma\sqrt{\frac{\pi}{2}\ln n}\) growth. Interestingly, this expression is broadly consistent with the result derived in \cite{biroli2024exact}, as for sufficiently large values of \((n-k)\), we have \(H_{n-k} \approx \ln (n-k)\).

By symmetry of the centered Gaussian law,
\begin{equation}\label{eq:gauss_symmetry}
  \mathbb{E}[X_{(k)}]
  = -\,\mathbb{E}[X_{(n-k+1)}],
  \quad\text{when }\mu=0,
\end{equation}
and more generally, a nonzero mean \(\mu\) simply shifts every order statistic by \(\mu\):
\begin{equation}\label{eq:gauss_shift}
  \mathbb{E}[X_{(k)}\mid\mu]
  = \mu \;+\;\mathbb{E}[X_{(k)}\mid\mu=0].
\end{equation}
Combining \eqref{eq:approx_Xk_gauss}–\eqref{eq:gauss_shift} with the definition
\(\displaystyle T_{\mathrm{shrink}}^{(p)}
  = \frac{X_{(n-p)}-X_{(p+1)}}{\,X_{(n-p+1)}-X_{(p)}\,}\), 
one obtains the Gaussian analog of \eqref{eq:shrinkage_ratio_uniform}:
\begin{equation}\label{eq:shrinkage_gauss}
  \mathbb{E}\bigl[T_{\mathrm{shrink}}^{(p)}\bigr]
  \;\approx\;
  \frac{1-\tfrac{H_p}{2\ln n}}{\,1-\tfrac{H_{p-1}}{2\ln n}\,}.
\end{equation}
This formula holds for any mean \(\mu\) and variance \(\sigma^2\), since both cancel in the ratio.   Figure~\ref{fig:diameter_shrinkage_statistics} compares the uniform prediction
\(\tfrac{n-2p-1}{n-2p+1}\) and the Gaussian approximation \eqref{eq:shrinkage_gauss} against Monte Carlo simulations.  In the Gaussian case we observe a small, nearly constant bias \(\alpha\approx0.03\).  To improve empirical fit, one may therefore use
\begin{equation}\label{eq:shrinkage_gauss_shifted}
  \mathbb{E}\bigl[T_{\mathrm{shrink}}^{(p)}\bigr]
  \;\approx\;
  \frac{1-\tfrac{H_p}{2\ln n}}{1-\tfrac{H_{p-1}}{2\ln n}} \;-\;\alpha.
\end{equation}
In the next section we leverage these analytical shrinkage curves to construct combined likelihood‐ and geometry‐based tests for distinguishing heavy‐tailed (Gaussian) from light‐tailed (uniform) point clouds.
%%%%%%%%%%%%%%%%%%%%%%%%%%%%%%%%%%%%%%%%%%%%%%%%
\subsection{Shift of the Sample Mean when Removing an Extreme Gaussian Outlier}
%%%%%%%%%%%%%%%%%%%%%%%%%%%%%%%%%%%%%%%%%%%%
We now quantify how the sample mean (center of mass) is shifted when we discard the single largest observation from \(n\) i.i.d.\ draws \(X_{(1)}\le\cdots\le X_{(n)}\sim\mathcal N(\mu,\sigma^2)\). By definition,
\[
  \bar X
  = \frac1n\sum_{i=1}^n X_{(i)},
  \quad
  \mathbb{E}[\bar X]
  = \frac1n\sum_{i=1}^n\mathbb{E}[X_{(i)}]
  = \mu,
\]
since the sample mean is unbiased.
%%%%%%%%%%%%%%%%%%%%%%%%%%%%%%%%%%%
\paragraph{Trimmed mean after removing the maximum.}
Delete \(X_{(n)}\) and form the \((n-1)\)-point mean
\(\displaystyle
  \bar X_{-}
  = \frac1{n-1}\sum_{i=1}^{n-1}X_{(i)}.
\)
Its expectation is
\[
  \mathbb{E}[\bar X_{-}]
  = \frac1{n-1}\sum_{i=1}^{n-1}\mathbb{E}[X_{(i)}].
\]
By symmetry of the centered Gaussian order‐statistic expectations (cf.\ \eqref{eq:gauss_symmetry}),
\(\sum_{i=2}^{n-1}\mathbb{E}[X_{(i)}]=(n-2)\mu\), and from \eqref{eq:approx_Xk_gauss} we have
\(\mathbb{E}[X_{(1)}]\approx\mu -\tfrac{\sigma\sqrt{2\pi\ln n}}2\).  Therefore
\[
  \mathbb{E}[\bar X_{-}]
  \approx
  \frac1{n-1}\Bigl[(n-2)\mu + \Bigl(\mu - \tfrac{\sigma\sqrt{2\pi\ln n}}2\Bigr)\Bigr]
  = \mu - \frac{\sigma\sqrt{2\pi\ln n}}{2(n-1)}.
\]
%%%%%%%%%%%%%%%%%%%%%%%%%%%%%%%%%%%
\paragraph{Net shift.}
The expected shift in the center of mass upon removal of the single largest point is thus
\[
  \mathbb{E}\bigl[\bar X - \bar X_{-}\bigr]
  \approx
  \frac{\sigma\sqrt{2\pi\,\ln n}}{2(n-1)}.
\]
 By an identical argument for the smallest point—and by symmetry—the simultaneous removal of both extremes produces no first‐order shift in the mean.
%%%%%%%%%%%%%%%%%%%%%%%%%%%%%%%%%%%%%%%%%%
\begin{figure*}[!h]
\centerline{\includegraphics[width=1\columnwidth]{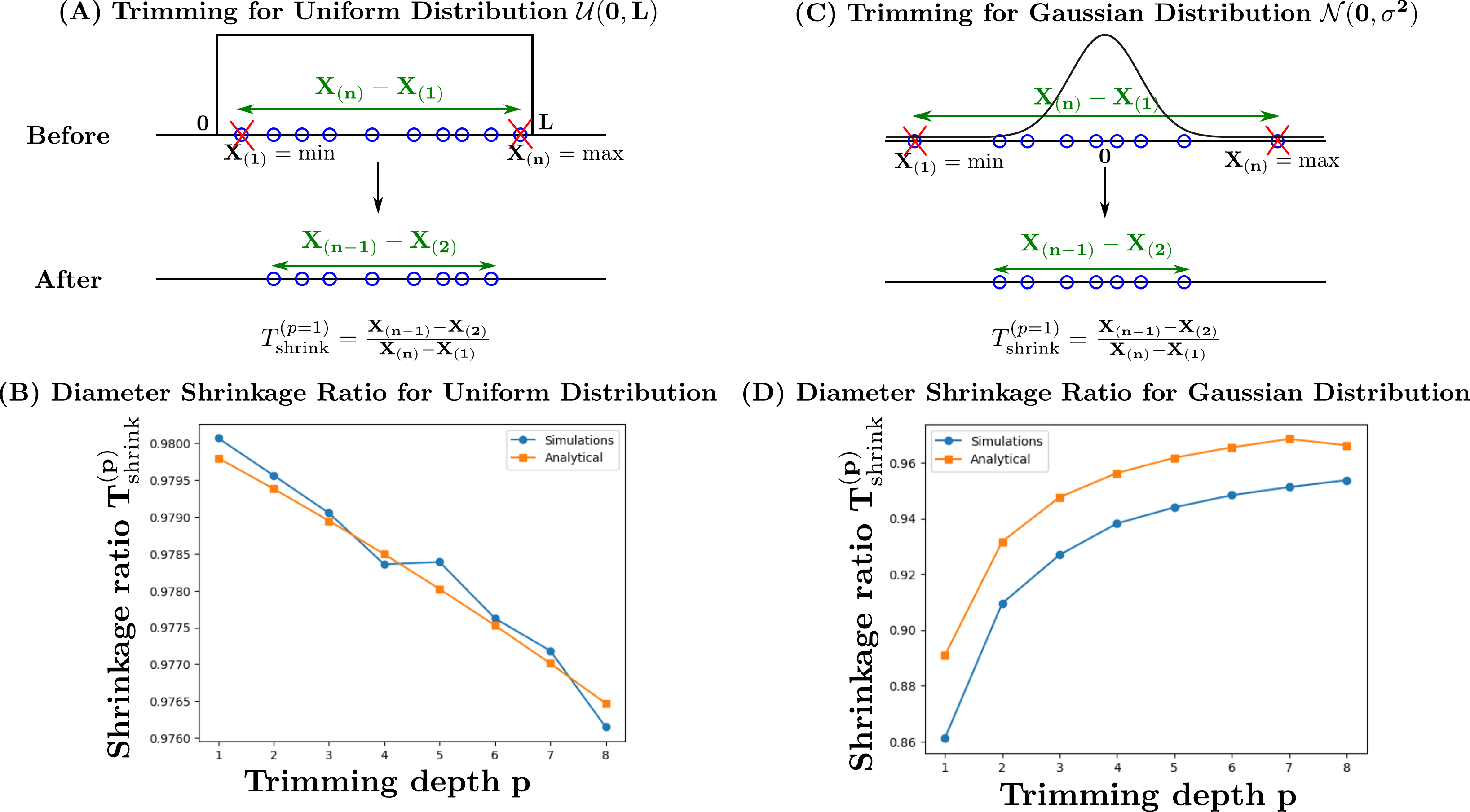}}
\caption{{\bf Diameter Shrinkage Statistics.} (A) and (B): Diameter Shrinkage Ratio after trimming for Uniform Distribution with $L = 1$. (C) and (D): Diameter Shrinkage Ratio after trimming for Gaussian Distribution with $\sigma=1$. For both distribution we used $n=100$ and the simulations are averaged over $10,000$ runs.}
\label{fig:diameter_shrinkage_statistics}
\end{figure*}
%%%%%%%%%%%%%%%%%%%%%%%%%%%%%%%%%%%%
%%%%%%%%%%%%%%%%%%%%%%%5
Fig. \ref{fig:diameter_shrinkage_statistics} panels (A)/(B) compare the uniform and Gaussian diameter‐shrinkage curves (with \(n=100\), averaged over \(10^4\) trials) and demonstrates the quality of our analytical approximation. In the next section, we integrate these geometric diagnostics into a robust clustering criterion.
%%%%%%%%%%%%%%%%%%%%%%%%%%%%%%%%%%%%%%
\section{Identification of meaningful clusters}
%%%%%%%%%%%%%%%%%%%%%%%%%%%%%%%%%%%%%%%%%%%%%%%%%%%%%%%
As stated in \autoref{sec:intro}, many spatial clustering algorithms operate under the assumption that clusters are simply regions of high point density separated by areas of lower density, without considering  what actually constitutes a meaningful cluster. Clustering methods such as K-means or DBSCAN typically rely on geometric heuristics (e.g.\ minimizing within‐cluster variance or finding dense neighborhoods) to partition data into groups. They search for geometric proximity or density thresholds, neglecting whether the points are close to each other by coincidence or due to an underlying distribution. However, these methods do not by themselves answer whether a detected “cluster’’ actually reflects an underlying generative mechanism, or is simply a random fluctuation. To address this, we propose a two‐statistical test that asks: \emph{does this one‐dimensional point cloud look more like a uniform distribution, or from a Gaussian distribution?}  Rejecting the uniform null in favor of a Gaussian alternative provides evidence of a true “hotspot’’ rather than mere spatial chance. As a result, classical algorithms may identify clusters that are mathematically valid but meaningless or misleading in practical applications.
%%%%%%%%%%%%%%%%%%%%%%%%%%%%%%5
\subsection{Two Competing Models}\label{sec:statistics_test}
%%%%%%%%%%%%%%%%%%%%%%%%%%%%%%%%%%%%%%
Starting with \(x_1,\dots,x_n\)  observed points on the real line, and write \(X_{(1)}\le\cdots\le X_{(n)}\) for their order statistics.  We compare:
\[
  \begin{aligned}
    H_0:~&x_i\;\overset{\text{i.i.d.}}{\sim}\;\mathcal U[a,b],
    &&\text{(unknown endpoints \(a,b\))},
    \\
    H_1:~&x_i\;\overset{\text{i.i.d.}}{\sim}\;\mathcal N(\mu,\sigma^2),
    &&\text{(unknown mean \(\mu\), variance \(\sigma^2\))}.
  \end{aligned}
\]
Two natural, complementary approaches are: \emph{Likelihood‐Based Model Selection}, vs \emph{Diameter‐Shrinkage–Based Test} that we shall present now.
%%%%%%%%%%%%%%%%%%%%%%%%%%%%%%%%%%%%%%%%%%%%%%
\subsubsection*{(a) Likelihood‐Based Model Selection}
%%%%%%%%%%%%%%%%%%%%%%%%%%%%%%%%%%%%%%%%%%%5
\paragraph{Uniform model.}  
Under \(\mathcal U[a,b]\), each observation has density \(\tfrac1{b-a}\) when \(x_i\in[a,b]\), and zero otherwise.  Thus the joint likelihood is
\[
  \mathcal L_U(a,b)
  = \prod_{i=1}^n \frac1{b-a}
  = (b-a)^{-n}
  \quad
  \text{provided }a\le X_{(1)}\le X_{(n)}\le b.
\]
Maximizing over \(a,b\) simply forces \(a=X_{(1)}\), \(b=X_{(n)}\), giving
\[
  \hat a = X_{(1)},\quad \hat b = X_{(n)},\qquad
  \ln\mathcal L_U = -\,n\,\ln\bigl(X_{(n)}-X_{(1)}\bigr).
\]
%%%%%%%%%%%%%%%%%%%%%%%%%%%%%%%%%%%%%%
\paragraph{Gaussian model.}  
For \(X_i\sim\mathcal N(\mu,\sigma^2)\), the likelihood is
\[
  \mathcal L_G(\mu,\sigma^2)
  = \prod_{i=1}^n
    \frac{1}{\sqrt{2\pi\sigma^2}}
    \exp\!\Bigl(-\tfrac{(x_i-\mu)^2}{2\sigma^2}\Bigr).
\]
One shows the MLEs are the sample mean and variance,
\(\hat\mu=\bar x\), \(\hat\sigma^2=\tfrac1n\sum(x_i-\bar x)^2\), yielding
\[
  \ln\mathcal L_G
  = -\tfrac n2\ln(2\pi\hat\sigma^2)\;-\;\tfrac n2.
\]
%%%%%%%%%%%%%%%%%%%%%%%%%%%%%%%%%%%%%%
\paragraph{Model comparison.}  
Assuming equal prior weight on \(H_0\) and \(H_1\), the posterior odds reduce to the likelihood‐ratio test:
\[
  \Lambda
  = \frac{\mathcal L_U(\hat a,\hat b)}
       {\mathcal L_G(\hat\mu,\hat\sigma^2)}
  \gtrless 1.
\]
Equivalently, we can compare \(\ln\mathcal L_U\) vs.\ \(\ln\mathcal L_G\) using the classical rule: assume a uniform prior over models, the posterior probability of the uniform model is:
\[
P(H_0 \mid \text{data}) = \frac{\mathcal{L}_U}{\mathcal{L}_U + \mathcal{L}_G}, \quad P(H_1 \mid \text{data}) = 1 - P(H_0 \mid \text{data}).
\]
The decision rule is straightforward:
\begin{itemize}
    \item Prefer \( H_0 \) (uniform) if \( \ln \mathcal{L}_U > \ln \mathcal{L}_G \),
    \item Prefer \( H_1 \) (Gaussian) otherwise.
\end{itemize}
%%%%%%%%%%%%%%%%%%%%%%%%%%%%%%%%%%%%%%
\subsubsection*{(b) Diameter‐Shrinkage–Based Test}
%%%%%%%%%%%%%%%%%%%%%%%%%%%%%%%%%%%%%%
As an alternative, model‐free indicator of tail behavior, we track how the sample diameter
\[
  D_p = X_{(n-p)} - X_{(p+1)}
\]
contracts as we trim \(p\) extreme points from each end.  Define the successive shrinkage ratio
\[
  T_{\mathrm{shrink}}^{(p)}
  = \frac{D_p}{D_{p-1}},
  \quad 1\le p<\tfrac n2.
\]
Under the uniform model one shows
\(\mathbb{E}[T_{\mathrm{shrink}}^{(p)}]\approx\tfrac{n-2p-1}{\,n-2p+1\,}\), 
whereas for a Gaussian sample
\(\displaystyle\mathbb{E}[T_{\mathrm{shrink}}^{(p)}]\approx
  \frac{1-\tfrac{H_p}{2\ln n}}{1-\tfrac{H_{p-1}}{2\ln n}} \)
(see Section~2.3). We derived the approximation:
\beq
\mathbb{E}(T_{\text{shrink}}^{(p)}) \approx \frac{1-\frac{H_p}{2\ln n}}{1-\frac{H_{p-1}}{2\ln n}} - \alpha
\eeq
The resulting shrinkage profile exhibits a slight increase with $p$, which can be attributed to the heavy tails of the Gaussian distribution. In practical applications, these shrinkage patterns provide a more informative characterization of the underlying distribution than a shrinkage computed at a fixed trimming depth $p$. In practice, we recommend computing \( T_{\text{shrink}}^{(p)} \) for several small values of \( p \) and comparing the empirical decay pattern against the theoretical curves for each model. This method can be particularly informative when the sample size is moderate and likelihood-based estimates are less stable. We then used this Diameter Shrinkage Statistics to construct a criterion for distinguishing between the Uniform and Gaussian distributions.  We shall now summarize our decision criteria. 
%%%%%%%%%%%%%%%%%%%%%%%%%%5
\subsubsection{Decision Rule and Geometric Classifier}
%%%%%%%%%%%%%%%%%%%%%%%%%%
We fix here a trimming depth \(p\ge1\) and the observed shrinkage ratios up to step \(p\) is defined as
\(\mathbf T_{\mathrm{emp}}^{(p)} = \bigl(T_{\mathrm{shrink}}^{(1)},\dots,T_{\mathrm{shrink}}^{(p)}\bigr)\). The expectation vectors \(\mathbf T_{U}^{(p)}\) and \(\mathbf T_{G}^{(p)}\) under \(H_0\) (uniform) and \(H_1\) (Gaussian), respectively, will be used to classify the sample by comparing Euclidean distances:
\[
  \phi_p(x_1,\dots,x_n)
  = 
  \begin{cases}
    1, & 
      \bigl\|\mathbf T_{\mathrm{emp}}^{(p)} - \mathbf T_{G}^{(p)}\bigr\|_2 
      < 
      \bigl\|\mathbf T_{\mathrm{emp}}^{(p)} - \mathbf T_{U}^{(p)}\bigr\|_2,
    \\[6pt]
    0, & \text{otherwise},
  \end{cases}
\]
where \(\phi_p=1\) denotes classification as Gaussian.   Figure~\ref{fig:diameter_shrinkage_test_procedure}(A–B) illustrates this geometric decision rule.
%%%%%%%%%%%%%5
\paragraph{Monte Carlo Calibration of \(p\)}
%%%%%%%%%%%%%%%%%%%%%%%%%%
To choose an optimal trimming depth \(p\), we performed the following simulations: for each \(p=1,2,\dots,15\), we generated \(N=1000\) samples of size \(n=100\) under both
\[
  H_0:\;x_i\sim\mathcal U[0,1],
  \quad
  H_1:\;x_i\sim\mathcal N(0,1).
\]
For each sample, we computed \(\mathbf T_{\mathrm{emp}}^{(p)}\), applied the rule above, and recorded the classification decision \(\phi_p\).  The empirical accuracy is computed as
\[
  \widehat{\mathrm{Accuracy}}(p)
  = \frac1N \sum_{j=1}^N \mathbf{1}\{\phi_p(\mathbf x^{(j)}) = y^{(j)}\},
\]
where \(y^{(j)}\in\{0,1\}\) is the true model label for the \(j\)th sample. Figure~\ref{fig:diameter_shrinkage_test_procedure}(C) shows  \(\widehat{\mathrm{Accuracy}}(p)\) vs \(p\).  Accuracy rises steadily with increasing \(p\), reaching a plateau around \(p=6\).  Beyond this point, further trimming reduces the number of remaining points too sharply, degrading performance.  Thus \(p=6\) offers an optimal trade–off between sensitivity to tail behavior and statistical sample size.
%%%%%%%%%%%%%%%%%%%%%%%%%%%%%%%55
\begin{figure*}[!h]
\centerline{\includegraphics[width=\textwidth]{images/diameter_shrinkage_test_procedure.pdf}}
\caption{Diameter‐shrinkage test procedure.  
(A) Pipeline: compute shrinkage ratios \(T_{\mathrm{shrink}}^{(1)},\dots,T_{\mathrm{shrink}}^{(p)}\) with \eqref{eq:def_shrinkage_step_p}.   
(B) Geometric decision: compare empirical ratios to the uniform and Gaussian expectation curves computed from \eqref{eq:shrinkage_gauss_shifted}
and \eqref{eq:shrinkage_ratio_uniform}.  
(C) Monte Carlo calibration: classification accuracy versus trimming depth \(p\) for \(N=1000\) samples of size \(n=100\).  
(D) Accuracy versus sample size \(n\), showing relative strengths of shrinkage‐based and likelihood‐ratio tests.}
\label{fig:diameter_shrinkage_test_procedure}
\end{figure*}
%%%%%%%%%%%%%%%%%%%%%%%%%%%%%%%
To conclude, we adopt \(p=6\) in subsequent applications.  The full algorithm, including a natural confidence measure based on relative distances to the two model curves, is given in the description of Algorithm~\ref{algo:diameter_shrinkage_statistics_algo} below: 
%%%%%%%%%%%%%%%%%%%%%%%%%%%%%%%%%%%
%%%%%%%%%%%%%%%%%%%%%%%%%%%%%%%%%%%%%%%%%%%%%%%5
\begin{algorithm}[h!]
\caption{DiameterShrinkageStatisticsDecision(\(P\), \(p=6\))}
\label{algo:diameter_shrinkage_statistics_algo}
\begin{algorithmic}[1]  
\State \textbf{Input:} 
  \begin{itemize}
    \item \(P\): set of \(n\) real‑valued points
    \item \(p\): trimming depth (default \(6\))
  \end{itemize}
\State \textbf{Compute empirical shrinkages:}
  \State Sort \(P\) to obtain order statistics \(\{X_{(1)}\le\cdots\le X_{(n)}\}\).
  \State Let \(D_0 = X_{(n)} - X_{(1)}\).
  \For{\(i=1\) to \(p\)}
    \State \(D_i \gets X_{(n-i)} - X_{(i+1)}\)
    \State \(T_{\mathrm{emp}}^{(i)} \gets D_i / D_{i-1}\)
  \EndFor
  \State Set \(\mathbf T_{\mathrm{emp}}^{(p)} = [T_{\mathrm{emp}}^{(1)},\dots,T_{\mathrm{emp}}^{(p)}]\).
\State \textbf{Compute theoretical curves:}
  \State \(\mathbf T_U^{(p)}\) via \eqref{eq:shrinkage_ratio_uniform}.
  \State \(\mathbf T_G^{(p)}\) via \eqref{eq:shrinkage_gauss_shifted}.
\State \(\quad d_U \gets \|\mathbf T_U^{(p)} - \mathbf T_{\mathrm{emp}}^{(p)}\|_2,\quad
       d_G \gets \|\mathbf T_G^{(p)} - \mathbf T_{\mathrm{emp}}^{(p)}\|_2\).
\If{\(d_G < d_U\)}
  \State \textbf{Output:} Gaussian, confidence \(1 - d_G/(d_G + d_U)\).
\Else
  \State \textbf{Output:} Uniform, confidence \(1 - d_U/(d_G + d_U)\).
\EndIf
\end{algorithmic}
\end{algorithm}

%%%%%%%%%%%%%%%%%%%%%%%%%%%%%%%%%%%%%%%%%%%%%%%
\subsubsection{Empirical Comparison of Likelihood‐Ratio and Shrinkage‐Based Classifiers}
%%%%%%%%%%%%%%%%%%%%%%%%%%%%%%%%%%%%%%%%%%%%%%%
To assess the relative robustness of the two approaches, we use a Monte Carlo approach: for each sample size 
\[
  n \in \{15,20,25,30,40,50,60,70,80,90,100,150,200\},
\]
we generated:
\begin{itemize}
  \item \(N=100\) independent samples of size \(n\) from \(\mathcal U(0,L)\), with the interval length \(L\) drawn uniformly from \([0,20]\);
  \item \(N=100\) independent samples of size \(n\) from \(\mathcal N(0,\sigma^2)\), with the standard deviation \(\sigma\) drawn uniformly from \([0,20]\).
\end{itemize}
Table~\ref{tab:comparison_methods} summarizes the overall classification accuracy and AUC (ROC) achieved by each method:
%%%%%%%%%%%%%%%%%%%%%%%%%%%%%%%%%%%%%%%%%%%
\begin{table}[ht]
\centering
\begin{tabular}{|l|c|c|}
\hline
\textbf{Method} & \textbf{Mean Accuracy} & \textbf{AUC (ROC)} \\
\hline
MLE (likelihood‐ratio) & 0.851 & 0.851 \\
Shrinkage (\(T_{\mathrm{shrink}}^{(i\le p)},\,p=6\)) & 0.864 & 0.864 \\
\hline
\end{tabular}
\caption{Aggregate performance of the MLE versus shrinkage‐based classifiers across varying sample sizes.}
\label{tab:comparison_methods}
\end{table}
%%%%%%%%%%%%%%%%%%%%%%%%%%%%%%%%
Figure~\ref{fig:diameter_shrinkage_test_procedure}D shows the classification accuracy vs \(n\).  The shrinkage‐based classifier outperforms the likelihood‐ratio test for moderate sample sizes (\(20\le n\le60\)), where our tail‐trimming approximations remain valid but likelihood estimates are still noisy.  For very small \(n<20\), neither method is reliable; for large \(n>60\), the MLE becomes nearly optimal as the data volume overwhelms the impact of trimming.\\
A possible interpretation of those results is that if $n<20$, the data is too scarce to be trimmed and the sample size is too low for our approximations to be valid. On the other hand, when $n > 60$, the data are large enough for the likelihood to account for the extreme points underlying distribution. To summarize, to classify i.i.d. real-valued samples \( x_1, \dots, x_n \in \mathbb{R} \) into a Uniform model or a Gaussian model, depending on the sample size \( n \), we recommend the following decision rules:
%%%%%%%%%%%%%%%%%%%%%%%%%%%%%%5
\begin{itemize}
  \item \emph{Moderate sample sizes} (\(20<n<60\)): employ the shrinkage‐based test, which additionally provides a natural confidence metric via distance to the theoretical shrinkage curve.
  \item \emph{Other regimes} (\(n\le20\) or \(n>60\)): default to the classical likelihood‐ratio test, leveraging its asymptotic efficiency.
\end{itemize}
In the next section, we demonstrate how this hybrid decision rule effectively identifies meaningful clusters in real‐world one‐dimensional spatial data.
%%%%%%%%%%%%%%%%%%%%%%%%%%%%%%%%%%%%%%%%%%%%%%%
\subsection{Validation of One-Dimensional Clustering shrinkage method} \label{sec:application_1d}
\begin{figure*}[http!]
\centerline{\includegraphics[width=1\columnwidth]{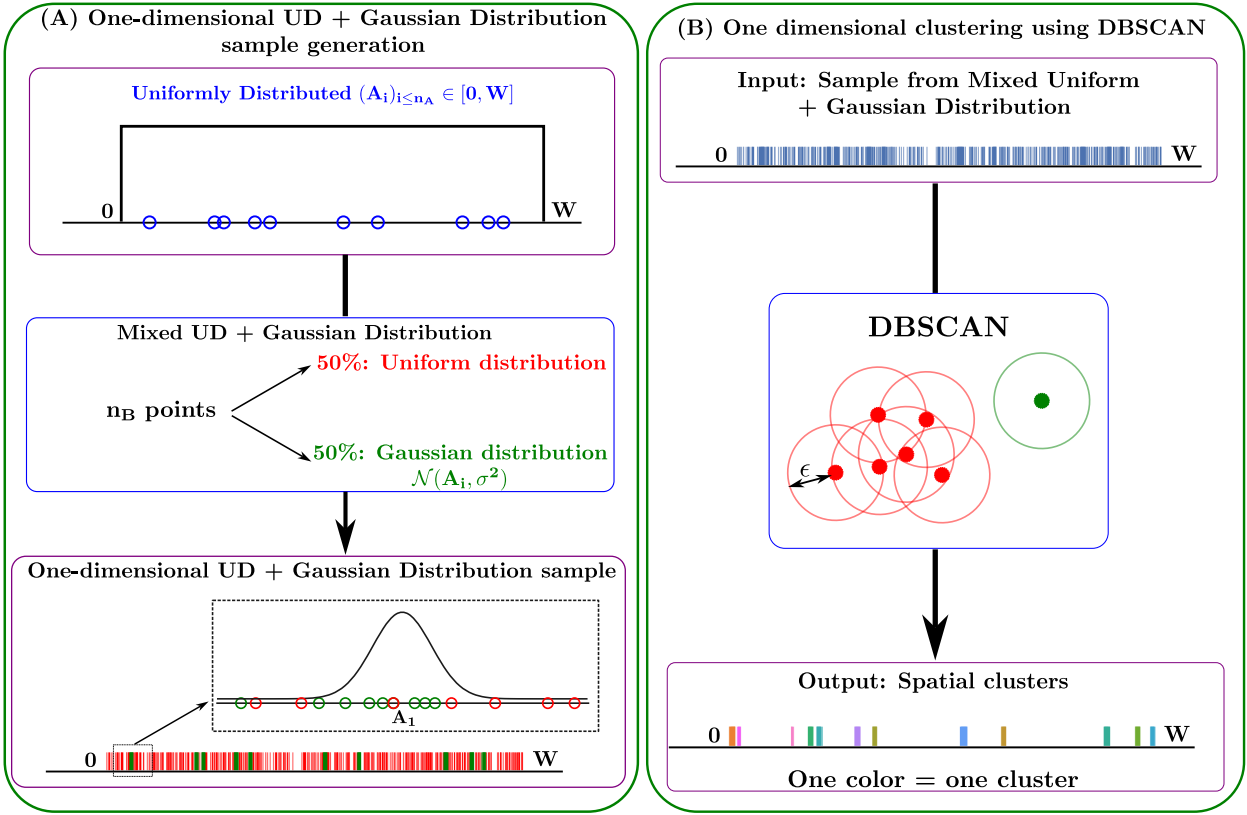}}
\caption{Generation of one-dimensional spatial clusters. (A): Generation of one-dimensional sample following a mixed Uniform and Gaussian Distribution on a segment of width $W=10,000$. Among the $n_B=1000$ points following the mixed UD + Gaussian Distribution, $50\%$ of them are drawn from a Gaussian distribution located around one of the $n_A=10$ anchor points with a variance of $\sigma^2=20$. (B): Application of DBSCAN algorithm, with parameters \(\varepsilon=20\) and \(\mathrm{min\_{samples}}=7\), on a one-dimensional sample following a mixed Uniform + Gaussian Distribution, yielding spatial clusters. }
\label{fig:ground_truth_dataset_clusters}
\end{figure*}

%%%%%%%%%%%%%%%%%%%%%%%%%%%%%%%%%%%%%%%%%%%%%%%
We assess here our method’s ability to identify statistically meaningful clusters in one dimension by applying it to data ( \(N=100\) independent datasets) generated as follows (see Fig.~\ref{fig:ground_truth_dataset_clusters}-(A):
\begin{enumerate}
  \item Sample \(n_A=10\) “anchor’’ locations uniformly on the interval \([0, W]\) with \(W=10{,}000\).
  \item Generate \(n_B=1000\) additional points, of which
    \begin{itemize}
      \item 50\% are drawn from \(\mathcal{N}(\mu_i,\sigma^2)\) with \(\sigma=20\), where each \(\mu_i\) is chosen uniformly from the \(n_A\) anchor set,
      \item 50\% are drawn uniformly from \([0,W]\) as background noise.
    \end{itemize}
\end{enumerate}
Each dataset was clustered with DBSCAN (Fig.~\ref{fig:ground_truth_dataset_clusters}-(B) using \(\varepsilon=20\) and \(\mathrm{min\_{samples}}=7\), parameters chosen to recover the visually apparent Gaussian clusters.  A DBSCAN cluster was labeled “significant’’ if more than half of its points originated from the Gaussian component. For each significant cluster, we then applied  
\begin{itemize}
  \item the combined MLE–shrinkage classifier (Fig.~\ref{fig:diameter_shrinkage_test_procedure}), and 
  \item the classical likelihood‐ratio test alone.
\end{itemize}
Over the \(N=100\) trials, the hybrid MLE–shrinkage method achieved a balanced accuracy of 0.943, versus 0.920 for the likelihood‐ratio test.  This improvement underscores the added robustness and discriminative power conferred by the diameter‐shrinkage statistic—especially notable given that it requires no density fitting or parameter tuning.
%%%%%%%%%%%%%%%%%%%%%%%%%%%%%%%%%%%%%%%
%%%%%%%%%%%%%%%%%%%%%%%%%%%%%%%%%%%%%%%
%%%%%%%%%%%%%%%%%%%%%%%%%%%%%%%%%%%%%%%
\section{Conclusion}
%%%%%%%%%%%%%%%%%%%%%%%%%%%%%%%%%%%%%%%
We have introduced a novel geometric test statistic, based on successive diameter shrinkage, for discriminating between light‐tailed (uniform) and heavy‐tailed (Gaussian) point cloud clustering in one dimension.  In contrast to the classical likelihood‐ratio test—which is asymptotically efficient under correct model specification but can suffer in small samples or in the presence of outliers—our shrinkage‐based criterion directly exploits the extreme‐value geometry of the data. 
This delivers superior discrimination in small‐sample or noisy settings.  By monitoring how the sample’s span contracts under successive removal of outliers, it exposes tail behavior without ever fitting a full density. In addition, we derived  here analytic approximations for the expected shrinkage under both uniform and Gaussian hypotheses, and demonstrated that these curves separate cleanly even for moderate sample sizes. The resulting procedure has several properties:
%%%%%%%%%%%%%%%%%%%%%%
\begin{itemize}
  \item {\bf Robustness.}  By focusing on the contraction of the sample span under trimming, the method remains stable under moderate deviations from the assumed density and can account robustly for individual outliers.
  \item {\bf Nonparametric character.}  No explicit density estimation is required; only the order statistics are needed to compute the test statistic.
  \item {\bf Extendibility.}  The same geometric principle may be generalized to higher dimensions via convex‐hull shrinkage or random projections, opening the door to robust cluster validation in multivariate settings.
\end{itemize}
We used Monte Carlo simulations to confirm that the diameter‐shrinkage test complements maximum‐likelihood methods, achieving superior classification accuracy in small‐sample and noisy regimes while retaining consistency in large‐sample limits.  In practice, the two approaches are highly complementary: The MLE‐based likelihood‐ratio test excels when the sample size is large and the model assumptions hold exactly, whereas the shrinkage‐based test can be used when robustness to outliers or moderate deviations is critical. Moreover, the shrinkage framework can generalize to higher dimensions—via convex‐hull trimming, random projections, or other geometric summaries—offering a toolkit for cluster validation and distributional inference. In future work, it would be of interest to study the properties of the statistic in \eqref{eq:def_shrinkage_step_p} directly, without relying on the approximation in \eqref{eq:expectation_approx}, particularly in the case of the Gaussian distribution.
%%%%%%%%%%%%%%%%%%%%%%%%%%%%%%%%%%%%%%%
\newpage
%%%%%%%%%%%%%%%%%%%%%%%%%%%%%%%%%
\section*{Appendix A. Order‐Statistic Perturbations under Uniform Sampling} \label{sec:appendix_uniform}
%%%%%%%%%%%%%%%%%%%%%%%%%%%%%%%%%%%%%%%%%%%%%%%
In this appendix we derive the exact distribution of the order statistics of \(n\) i.i.d.\ uniform \([0,L]\) samples, and then compute how their span (diameter) and centre‐of‐mass shift when the most extreme points are removed.
%%%%%%%%%%%%%%%%%%%%%%%%%%%%%%%%55
\paragraph{Joint density of ordered samples.}  
%%%%%%%%%%%%%%%%%%%%%%%%%%
Since each of the \(n!\) permutations of \((X_{(1)},\dots,X_{(n)})\) is equally likely and each \(X_i\) has density \(1/L\) on \([0,L]\), the joint density of the sorted vector \((X_{(1)},\dots,X_{(n)})\) is simply
\[
  f(x_1,\dots,x_n)
  = \frac{n!}{L^n}\,
    \mathbf1_{\{0\le x_1\le\cdots\le x_n\le L\}}.
\]
Outside the simplex \(0\le x_1\le\cdots\le x_n\le L\) the density vanishes, and inside it is constant.
%%%%%%%%%%%%%%%%%%%%%%%%%%
\subsection*{A.1. Marginal density of the \(k\)th order statistic}
%%%%%%%%%%%%%%%%%%%%%%%%%%
To isolate the \(k\)th statistic \(X_{(k)}\), we integrate out all other coordinates.  One finds by repeated integration of polynomials that
\[
  f_k(x)
  = \int_{\substack{0\le x_1\le\cdots\le x_{k-1}\le x \\[1pt] x\le x_{k+1}\le\cdots\le x_n\le L}}
    \frac{n!}{L^n}
    \,dx_1\cdots dx_{k-1}\,dx_{k+1}\cdots dx_n
  = \frac{n!}{(k-1)!\,(n-k)!}\,\frac{x^{k-1}(L-x)^{n-k}}{L^n},
\]
for \(0\le x\le L\).  Equivalently, \(X_{(k)}/L\) follows a \(\mathrm{Beta}(k,n-k+1)\) law. Indeed, a direct computation leads to 
\beq
f_k(x_k) & = \frac{n!}{L^n} \left(\int_0^{x_k} dx_1 \int_{x_1}^{x_k}dx_2 \dots  \int_{x_{k-2}}^{x_k}dx_{k-1}\right) \left(\int_{x_k}^L d_{x_{k+1}} \int_{x_{k+1}}^L d_{x_{k+2}} \dots \int_{x_{n-1}}^L dx_{n}\right) \\
&= \frac{n!}{L^n} L(x_k)  R(x_k),
\eeq
where
\beq 
L(x_k) = \int_0^{x_k} dx_1 \int_{x_1}^{x_k}dx_2 \dots  \int_{x_{k-2}}^{x_k}dx_{k-1}\\
R(x_k) = \int_{x_k}^L d_{x_{k+1}} \int_{x_{k+1}}^L d_{x_{k+2}} \dots \int_{x_{n-1}}^L dx_{n}
\eeq 
Using that for  $a, b \in \mathbb{R}$ and $N \in \mathbb{N}$, we have \beq 
\int_a^b (b-x)^Ndx = \frac{1}{N+1}(b-a)^{N+1}
\eeq
A direct integration leads to 
\beq 
L(x_k) = \int_0^{x_k} dx_1 \int_{x_1}^{x_k}dx_2 \dots  \int_{x_{k-3}}^{x_k}(x_k - x_{k-2})dx_{k-2} = \frac{x_k^{k-1}}{(k-1)!} 
\eeq
Similarly, 
\beq 
R(x_k) = \frac{(L-x_k)^{n-k}}{(n-k)!}.
\eeq
Finally, the distribution of $X_{(k)}$ is given by
\beq\label{eq:distribution_xkp}
{f_k(x_k) = \frac{n!}{L^n} \frac{x_k^{k-1}}{(k-1)!}\frac{(L-x_k)^{n-k}}{(n-k)!}}
\eeq
%%%%%%%%%%%%%%%%%%%%%%%%%%
\subsection*{A.2. Expected value and higher moments}
%%%%%%%%%%%%%%%%%%%%%%%%%%
The closed‐form density immediately yields moments. The first moment is be expressed as:
\beq 
\eE(X_{(k)}) = \int_0^L x_k f_k(x_k)dx_k
\eeq 
which can be computed by using  eq. \eqref{eq:distribution_xk} as 
\beq
\eE(X_{(k)}) = \frac{n!}{L^n (k-1)! (n-k)!}\int_0^L x_k^{k} (L-x_k)^{n-k} dx_k. \eeq 
Using the $\beta-$ function defined as:
\beq\label{eq:beta}
\beta(x,y)=\int_0^1 t^{x-1}(1-t)^{y-1}dt
\eeq
We get by using the substitution $t = \frac{x_k}{L}$, $dt = \frac{dx_k}{L}$ 
\beq 
    \eE(X_{(k)}) & = \frac{n!}{L^n (k-1)! (n-k)!}\int_0^1 L^{k}t^k L^{n-k}(1-t)^{n-k}L dt \\ 
    & = \frac{n!}{L^n (k-1)! (n-k)!} L^{n+1} \int_0^1 t^k (1-t)^{n-k}dt \\
    & = \frac{n!}{L^n (k-1)! (n-k)!} L^{n+1} \beta(k+1, n-k+1)
\eeq
Finally, using for $x,y \in \mathbb{R}$,
\beq \label{eq:beta_gamma}
\beta(x,y)=\frac{\Gamma(x)\Gamma(y)}{\Gamma(x+y)},
\eeq
where 
\beq \label{eq:gamma}
\Gamma(z) = \int_0^{+\infty}t^{z-1}e^{-t}dt \eeq
and for integer $n$, $\Gamma(n) = (n-1)!$,  we obtain 
\beq
    \eE(X_{(k)}) & = \frac{n!}{L^n (k-1)! (n-k)!} L^{n+1} \frac{\Gamma(k+1)\Gamma(n-k+1)}{\Gamma(n+2)} \\
 & = \frac{n!}{(k-1)! (n-k)!} L \frac{(k)!(n-k)!}{(n+1)!}.
\eeq 
Finally, the expected value of $X_{(k)}$ can be expressed as:
\beq\label{eq:esperance_xk}
\eE(X_{(k)}) = L \frac{k}{n+1}.
\eeq
More generally, for integer \(p\ge1\),
\[
  \mathbb{E}[X_{(k)}^p]
  = \int_0^L x^p f_k(x)\,dx
  = L^p\,\frac{k\,(k+1)\cdots(k+p-1)}
               {(n+1)\,(n+2)\cdots(n+p)}.
\]

All these results and computations can be found in the literature \cite{karlin1981second, david2004order, majumdar2024statistics}.

%%%%%%%%%%%%%%%%%%%%%%%%%%
\subsection*{A.3. Trimmed diameter}
%%%%%%%%%%%%%%%%%%%%%%%%%%
Define the “\(p\)–trimmed diameter’’ by removing the \(p\) smallest and \(p\) largest points:
\[
  D_p \;=\;X_{(n-p)}-X_{(p+1)}.
\]
Since \(\mathbb{E}[X_{(m)}]=L\,\tfrac{m}{n+1}\), it follows that
\[
  \mathbb{E}[D_p]
  = \mathbb{E}[X_{(n-p)}]-\mathbb{E}[X_{(p+1)}]
  = L\,\frac{(n-p)-(p+1)}{n+1}
  = L\,\frac{n-2p-1}{n+1}.
\]

For $p = 0$, this result can also be obtained as follows: first, the joint distribution of $X_{(1)}, X_{(n)}$ is given by 
\beq 
f(x_1, x_n) = \int_0^L \dots  \int_0^L f(x_1, \dots, x_n) dx_2 \dots dx_{n-1}
\eeq 
By using \eqref{eq:density}, we have, if $0 \leq  x_1 \leq x_n \leq L$:
\beq
f(x_1, x_n) &= \frac{n!}{L^n}\int_{x_1}^{x_n} dx_2\dots  \int_{x_n-2}^{x_n} dx_{n-1} \\
& = \frac{n!}{L^n}\frac{(x_n - x_1)^{n-2}}{(n-2)!}
\eeq 
Hence, the joint distribution of $X_{(1)}, X_{(n)}$ can be expressed as:
\beq\label{eq:density}
  f(x_1, x_n) = \left\{
    \begin{array}{ll}
       \frac{n!}{L^n}\frac{(x_n - x_1)^{n-2}}{(n-2)!} & \mbox{if } 0 \leq x_1 \leq  x_n \leq L \\
        0 & \mbox{else}
    \end{array}
\right.
\eeq
Therefore,
\beq
        \eE(X_{(n)} - X_{(1)}) & = \int_0^L \int_0^L (x_n - x_1)f(x_1, x_n)dx_1dx_n
         = \int_0^L \int_{x_1}^L \frac{n!}{L^n}\frac{(x_n - x_1)^{n-1}}{(n-2)!}dx_ndx_1 \\
         &= \frac{n!}{L^n(n-2)!} \int_0^L \int_{x_1}^L (x_n - x_1)^{n-1}dx_ndx_1 
         = \frac{n!}{L^n(n-2)!} \int_0^L (-\int_L^{x_1} (x_n - x_1)^{n-1}dx_n)dx_1 \\
        & = \frac{n!}{L^n(n-2)!} \int_0^L (- \frac{1}{(-1)^{n-1}}\int_L^{x_1} (x_1 - x_n)^{n-1}dx_n)dx_1 = \frac{n!}{L^n(n-2)!} \int_0^L (-\frac{1}{(-1)^{n-1}}\frac{(x_1-L)^n}{n})dx_1 \\
        & = \frac{n!}{L^n(n-2)!} \int_0^L (-\frac{1}{(-1)^{n-1}(-1)^n}\frac{(L-x_1)^n}{n})dx_1 \\
        & = \frac{n!}{L^n(n-2)!} \int_0^L (\frac{1}{(-1)^{2n-2}}\frac{(L-x_1)^n}{n})dx_1 = \frac{n!}{L^n(n-2)!} \int_0^L (\frac{(L-x_1)^n}{n})dx_1 
         = \frac{n!}{L^n(n-2)!} \frac{L^{n+1}}{n(n+1)} \\
        & =  \frac{n(n-1)L}{n(n+1)} 
         = \frac{L(n-1)}{n+1}.
\eeq
%%%%%%%%%%%%%%%%%%%%%%%%%%
% \subsection*{A.4. Shift of the sample mean under extreme removal}
% %%%%%%%%%%%%%%%%%%%%%%%%%%
% Finally, we show that removing the single largest observation \(X_{(n)}\) shifts the sample mean by
% \(\tfrac{L}{2(n+1)}\) in expectation.  Writing the original mean as \(\bar X =\tfrac1n\sum_iX_i\) and the trimmed mean as \(\bar X_{\mathrm{trim}}=\tfrac1{n-1}\sum_{i=1}^{n-1}X_{(i)}\), we obtain
% \[
%   \mathbb{E}[\bar X-\bar X_{\mathrm{trim}}]
%   = \frac{L}{2(n+1)}.
% \]
% By symmetry, if one removes both the smallest and largest points, the expected shift of the mean vanishes.
%%%%%%%%%%%%%%%%%%%%%%%%%%%%%%%%%%%%%%5
\subsection*{A.4. Shrinkage ratio after trimming}
%%%%%%%%%%%%%%%%%%%%%%%%%%%%%%%%%%%%%%%%%
We compute here the expected value of the random variable 
\beq 
\frac{X_{n-1} - X_{(1)}}{X_{(n)} - X_{(1)}}
\eeq 
As was done previously, the first step is to find the joint distribution of $X_{(1)}, X_{n-1}, X_{(n)}$.
\beq
f(x_1, x_{n-1}, x_n) = \int_0^L \dots  \int_0^L f(x_1, \dots, x_n) dx_2 \dots dx_{n-2}
\eeq 
By using \eqref{eq:density}, we have, if $0 \leq  x_1 \leq x_{n-1} \leq x_n \leq L$:
\beq 
f(x_1, x_{n-1}, x_n)  = \frac{n!}{L^n} \int_{x_1}^{x_{n-1}} dx_2 \dots  \int_{x_{n-2}}^{x_{n-1}} dx_{n-3} =  \frac{n!}{L^n} \frac{(x_{n-1} - x_1)^{n-3}}{(n-3)!}
\eeq 
Hence, the joint distribution of $X_{(1)}, X_{(n)}$ can be expressed as:
\beq\label{eq:densityp}
  f(x_1, x_{n-1}, x_n) = \left\{
    \begin{array}{ll}
       \frac{n!}{L^n} \frac{(x_{n-1} - x_1)^{n-3}}{(n-3)!} & \mbox{if } 0 \leq  x_1 \leq x_{n-1} \leq x_n \leq L \\
        0 & \mbox{else}
    \end{array}
\right.
\eeq
Thus,
\beq \label{eq:intermediate_esperenace_ratio}
\eE(\frac{X_{n-1} - X_{(1)}}{X_{(n)} - X_{(1)}}) & = \int_0^L \int_0^L \int_0^L \frac{x_{n-1} - x_1}{x_n - x_1}f(x_1, x_{n-1}, x_n)dx_1dx_{n-1}dx_n \\
& = \frac{n!}{L^n} \int_0^L \int_{x_1}^L  \int_{x_{n-1}}^L 
\frac{x_{n-1} - x_1}{x_n - x_1} \frac{(x_{n-1} - x_1)^{n-3}}{(n-3)!} dx_n dx_{n-1} dx_1 \\
& = \frac{n!}{L^n(n-3)!} \int_0^L \int_{x_1}^L  \int_{x_{n-1}}^L  \frac{1}{x_n - x_1} (x_{n-1} - x_1)^{n-2} dx_n dx_{n-1} dx_1.
\eeq 
We shall use the notation
\beq
I = \int_0^L \int_{y}^L  \int_{z}^L  \frac{(z-y)^{n-2}}{x-y}dxdzdy
\eeq 
First, we consider 
\beq 
    I_1 & = \int_{z}^L  \frac{1}{x-y} (z-y)^{n-2} dx = (z-y)^{n-2}\int_{z}^L \frac{1}{x-y}dx \\
    & = (z-y)^{n-2} \left( \ln(L-y) -\ln(z-y)\right)
\eeq 
Then,
\beq
    I_2 & = \int_{y}^L(z-y)^{n-2} \left( \ln(L-y) -\approx \ln(z-y)\right)dz \\
    & = \int_{y}^L(z-y)^{n-2}\ln(L-y)dz - \int_{y}^L(z-y)^{n-2}\ln(z-y)dz
\eeq 
By integrating by parts:
\beq
    I_2 & = \ln(L-y)\int_{y}^L(z-y)^{n-2}dz - \left( \left[ \frac{1}{n-1}(z-y)^{n-1} ln(z-y)\right]_y^L - \int_{y}^L\frac{1}{n-1}(z-y)^{n-1}\frac{1}{z-y}dz \right) \\
    & = \ln(L-y)\frac{1}{n-1}(L-y)^{n-1} - \left( \left[ \frac{1}{n-1}(z-y)^{n-1} ln(z-y)\right]_y^L - \int_{y}^L\frac{1}{n-1}(z-y)^{n-1}\frac{1}{z-y}dz \right) \\
    & = \ln(L-y)\frac{1}{n-1}(L-y)^{n-1} - \left(\frac{1}{n-1}(L-y)^{n-1} ln(L-y) - 0 - \int_{y}^L\frac{1}{n-1}(z-y)^{n-2}dz \right) \\
    & = \ln(L-y)\frac{1}{n-1}(L-y)^{n-1} - \left(\frac{1}{n-1}(L-y)^{n-1} ln(L-y)- \frac{1}{(n-1)^2}(L-y)^{n-1} \right) \\
    & = \frac{1}{(n-1)^2}(L-y)^{n-1}
\eeq 
Finally,
\beq \label{eq:integral_complete}
    I = \int_0^L  \frac{1}{(n-1)^2}(L-y)^{n-1}dy = \frac{1}{(n-1)^2}\frac{L^n}{n}.
\eeq 
By merging \eqref{eq:integral_complete} and \eqref{eq:intermediate_esperenace_ratio}, we have:
\beq 
    \eE(\frac{X_{n-1} - X_{(1)}}{X_{(n)} - X_{(1)}})  = \frac{n!}{L^n(n-3)!}  \frac{1}{(n-1)^2}\frac{L^n}{n}   = \frac{n!}{L^n(n-3)!}  \frac{1}{(n-1)^2}\frac{L^n}{n} 
 = \frac{n-2}{n-1}
\eeq
Therefore, the expected value of segment length evolution ratio after removing of $X_{(n)}$ can be expressed by:
\beq\label{eq:esperance_evolution_ratio_segment_length_xn}
{\eE(\frac{X_{n-1} - X_{(1)}}{X_{(n)} - X_{(1)}}) = \frac{n-2}{n-1}}
\eeq
It is interesting to further notice that
\beq\label{eq:esperance_evolution_ratio_segment_length_xn_justify_approx}
    \eE(\frac{X_{n-1} - X_{(1)}}{X_{(n)} - X_{(1)}}) = \frac{n-2}{n-1} = \frac{\eE(X_{n-1} - X_{(1)})}{\eE(X_{n} - X_{(1)})},
\eeq
which somehow justifies the approximation \eqref{eq:expectation_approx}, stating that:
\[\mathbb{E}(\frac{D_p}{D_{p-1}}) \approx
  \frac{\mathbb{E}[D_p]}{\mathbb{E}[D_{p-1}]}
  = \frac{n-2p-1}{n-2p+1},
  \quad 1\le p<\tfrac n2.
\]

\section*{Appendix B: Alternative Elementary Derivation of \(\mathbb{E}[\max_{1\le i\le n}|S_i|]\) Without the Chu–Tanner Fit} \label{sec:alt_distribution_max_abs_gaussian}
%%%%%%%%%%%%%%%%%%%%%55
We recall that $X_1,...,X_n \stackrel{iid}{\sim} \mathcal{N}(0,\sigma^2)$ are defined by
$ M_n = \max_{1\leq i\leq n} |X_i|$.
%%%%%%%%%%%%%%%%%
We shall show here that the expected maximum satisfies:
\[ \mathbb{E}[M_n] = \sigma\sqrt{2\log n} + \frac{\sigma(\log(4\pi))}{2\sqrt{2\log n}} + o\left(\frac{1}{\sqrt{\log n}}\right) \]
where $\gamma$ is Euler's constant.
%%%%%%%%%%%%%%%%%%%%%%%%%5
The expectation decomposes as:
\[ \mathbb{E}[M_n] = \underbrace{\int_0^{a_n} \mathbb{P}(M_n > r)dr}_{I} + \underbrace{\int_{a_n}^\infty \mathbb{P}(M_n > r)dr}_{II} \]
where $a_n = \sigma\sqrt{2\log n}$. We shall now decompose the integral. This follows from:
\beq
I &= \int_0^{a_n} \mathbb{P}(M_n > r),dr 
= \int_0^{a_n} \left(1 - \mathbb{P}(M_n \leq r)\right)dr = \int_0^{a_n} 1,dr - \int_0^{a_n} \mathbb{P}(M_n \leq r),dr
\\
&= a_n - \int_0^{a_n} \left[\text{erf}\left(\frac{r}{\sigma\sqrt{2}}\right)\right]^n dr = a_n - J_n
\eeq
To study $J_n$, we use  \( \epsilon_n = \frac{\sigma\log\log n}{\sqrt{2\log n}} \) and split the integral \( J_n \) into:
\[
J_n = \underbrace{\int_0^{a_n - \epsilon_n} \text{erf}(\cdots)^n dr}_{J_n^{(1)}} + \underbrace{\int_{a_n - \epsilon_n}^{a_n} \text{erf}(\cdots)^n dr}_{J_n^{(2)}}
\]
To analyse  \( J_n^{(1)} \) in th Bulk Region), we have for \( r \leq a_n - \epsilon_n \):
\[
\frac{r}{\sigma\sqrt{2}} \leq \sqrt{\log n} - \frac{\log\log n}{2\sqrt{\log n}}
\]
Using the erf asymptotic expansion:
\[
\text{erf}(x) \leq 1 - \frac{e^{-x^2}}{x\sqrt{\pi}} \leq 1 - \frac{e^{-(\sqrt{\log n} - \delta_n)^2}}{2\sqrt{\pi\log n}}
\]
where \( \delta_n = \frac{\log\log n}{2\sqrt{\log n}} \).
We expand the exponent:
\[
(\sqrt{\log n} - \delta_n)^2 = \log n - \log\log n + \frac{(\log\log n)^2}{4\log n}
\]
Thus:
\[
e^{-(\cdots)} = \frac{e^{\log\log n}}{n} \left(1 - \frac{(\log\log n)^2}{4\log n}\right) = \frac{\log n}{n}(1 + o(1))
\]
Therefore:
\[
\text{erf}\left(\frac{r}{\sigma\sqrt{2}}\right) \leq 1 - \frac{\log n}{2n\sqrt{\pi\log n}}(1 + o(1)) = 1 - \frac{\sqrt{\log n}}{2n\sqrt{\pi}}(1 + o(1))
\]
Raising to the \( n \)-th power:
\[
\left[1 - \frac{\sqrt{\log n}}{2n\sqrt{\pi}}(1 + o(1))\right]^n \leq \exp\left(-\frac{\sqrt{\log n}}{2\sqrt{\pi}}(1 + o(1))\right) \leq e^{-c\sqrt{\log n}}
\]
Thus:
\[
J_n^{(1)} \leq (a_n - \epsilon_n)e^{-c\sqrt{\log n}} \leq \sigma\sqrt{2\log n}\cdot e^{-c\sqrt{\log n}} \to 0 \text{ exponentially fast.}
\]
We now analyse $J_n^{(2)}$ (Boundary Layer): for $r \in [a_n - \epsilon_n, a_n]$, we get 
\[
r = a_n - \frac{\sigma t}{\sqrt{2\log n}}, \quad t \in [0, \log\log n]
\]
% The Jacobian is:
% \[
% dr = \frac{\sigma}{\sqrt{2\log n}} dt
% \]
The erf term becomes:
\[
\text{erf}\left(\sqrt{\log n} - \frac{t}{2\sqrt{\log n}}\right) \approx 1 - \frac{e^{-(\log n + t + t^2/(4\log n))}}{\sqrt{\pi}(\sqrt{\log n} - t/(2\sqrt{\log n}))}
\]
Simplifying:
\[
\approx 1 - \frac{e^{-t}}{n\sqrt{\pi\log n}}(1 + O(t/\log n))
\]
Thus:
\[
J_n^{(2)} \approx \frac{\sigma}{\sqrt{2\log n}} \int_0^{\log\log n} \left[1 - \frac{e^{-t}}{n\sqrt{\pi\log n}}(1 + O(t/\log n))\right]^n dt
\]
Using \( (1 - x/n)^n \approx e^{-x} \):
\[
\approx \frac{\sigma}{\sqrt{2\log n}} \int_0^{\log\log n} \exp\left(-\frac{e^{-t}}{\sqrt{\pi\log n}}\right) dt
\]
For large \( n \), this becomes:
\[
\approx \frac{\sigma}{\sqrt{2\log n}} \int_0^\infty \exp\left(-\frac{e^{-t}}{\sqrt{\pi\log n}}\right) dt = \frac{\sigma \gamma}{\sqrt{2\log n}}(1 + o(1))
\]
We get the final result:
\[
I = a_n - J_n^{(1)} - J_n^{(2)} = \sigma\sqrt{2\log n} - 0 - \frac{\sigma \gamma}{\sqrt{2\log n}} + o\left(\frac{1}{\sqrt{\log n}}\right)
\]
Thus:
\[
I = \sigma\sqrt{2\log n} - \frac{\sigma \gamma}{2\sqrt{2\log n}} + o\left(\frac{1}{\sqrt{\log n}}\right).
\]
We now analyze the tail contribution to the expected maximum:
\[
II = \int_{a_n}^\infty \mathbb{P}(M_n > r)\,dr
\]
where \( a_n = \sigma\sqrt{2\log n} \).
changing Variables
\[
r = a_n + \frac{\sigma y}{\sqrt{2\log n}}, \quad dr = \frac{\sigma}{\sqrt{2\log n}} dy
\]
The integral transforms to:
\[
II = \frac{\sigma}{\sqrt{2\log n}} \int_0^\infty \left[1 - \text{erf}\left(\sqrt{\log n} + \frac{y}{2\sqrt{\log n}}\right)^n\right] dy
\]
For \( x \gg 1 \):
\[
\text{erf}(x) = 1 - \frac{e^{-x^2}}{x\sqrt{\pi}} \left(1 - \frac{1}{2x^2} + O(x^{-4})\right)
\]
Let \( x = \sqrt{\log n} + \frac{y}{2\sqrt{\log n}} \). Then:
\[
x^2 = \log n + y + \frac{y^2}{4\log n} + O\left(\frac{y^3}{(\log n)^{3/2}}\right)
\]
Thus:
\[
e^{-x^2} = \frac{e^{-y}}{n} \left(1 - \frac{y^2}{4\log n} + O\left(\frac{y^3}{(\log n)^{3/2}}\right)\right)
\]
and:
\[
x\sqrt{\pi} = \sqrt{\pi\log n} \left(1 + \frac{y}{2\log n}\right)
\]
The survival function becomes:
\[
\mathbb{P}(M_n > r) = 1 - \left[1 - \frac{e^{-y}}{n\sqrt{\pi\log n}} \left(1 - \frac{y}{2\log n} - \frac{y^2}{4\log n} + O\left(\frac{y^3}{(\log n)^2}\right)\right)\right]^n
\]
Using \( (1 - \frac{c}{n})^n \approx e^{-c} \):
\[
\approx 1 - \exp\left(-\frac{e^{-y}}{\sqrt{\pi\log n}} \left(1 - \frac{y}{2\log n} - \frac{y^2}{4\log n}\right)\right)
\]
The integral decomposes as:
\[
II = \frac{\sigma}{\sqrt{2\log n}} \left[ \int_0^{\sqrt{\log n}} + \int_{\sqrt{\log n}}^\infty \right] \left[1 - \exp\left(-\frac{e^{-y}}{\sqrt{\pi\log n}}\right)\right] dy
\]

\textbf{Region 1:} \( 0 \leq y \leq \sqrt{\log n} \)
\[
1 - \exp\left(-\frac{e^{-y}}{\sqrt{\pi\log n}}\right) \approx \frac{e^{-y}}{\sqrt{\pi\log n}} - \frac{e^{-2y}}{2\pi\log n}
\]
Integrating:
\[
\int_0^{\sqrt{\log n}} \approx \frac{1 - e^{-\sqrt{\log n}}}{\sqrt{\pi\log n}} - \frac{1 - e^{-2\sqrt{\log n}}}{4\pi\log n}
\]

\textbf{Region 2:} \( y > \sqrt{\log n} \)
\[
\int_{\sqrt{\log n}}^\infty \approx \frac{e^{-\sqrt{\log n}}}{\sqrt{\pi\log n}}
\]
The dominant term comes from:
\[
\int_0^\infty \left[1 - \exp\left(-e^{-y}\right)\right] dy = \gamma + \log(4\pi)
\]
after appropriate rescaling. Combining all terms:
\[
II = \frac{\sigma}{\sqrt{2\log n}} \left[ (\gamma + \log(4\pi)) + O\left(\frac{1}{\sqrt{\log n}}\right) \right] = \frac{\sigma(\gamma + \log(4\pi))}{2\sqrt{2\log n}} + o\left(\frac{1}{\sqrt{\log n}}\right).
\]
%%%%%%%%%%%%%%%%%%%%%%%%%%%%%%%%%%%
\section*{Appendix C: Reshuffling for a Gaussian Distribution} \label{sec:AppendixGaussian}
%%%%%%%%%%%%%%%%%%%%%%%%%%%%%%%%%5
We recall here the distribution for $S_1, \dots, S_n$ i.i.d following a Gaussian law $\mathcal{N}(0, \sigma^2)$. Since there are $n!$ possible permutations and the density of a random variable following such Gaussian law is $\ds \frac{e^{-\frac{x^2}{2\sigma^2}}}{\sigma\sqrt{2 \pi}}dx$, the probability density function (pdf) of the order statistics $X_{(1)}, \dots, X_{(n)}$ is
\beq\label{eq:densitypp}
  f(x_1, \dots, x_n) = \left\{
    \begin{array}{ll}
       \ds \frac{n!}{(\sigma\sqrt{2 \pi})^n} e^{-\frac{\sum_{i=1}^n x_i^2}{2\sigma^2}}& \mbox{ for } x_1 \leq \dots \leq x_n  \\\\
        0 & \mbox{otherwise}
    \end{array}
\right.
\eeq
\subsection*{B.1. Expected value of $X_{(k)}$}
We now compute the mean associated to the $X_{(k)}-$variable. It can be expressed as 
\beq \label{eq:expected_value_integral}
\eE(X_{(k)}) = \int_{0}^{+\infty} (1 - F_k(x_k))dx_k - \int_{0}^{-\infty} F_k(x_k)dx_k,
\eeq 
where $F_k$ is the cumulative distribution function of the $X_{(k)}-$variable
\beq
F_k(x_k) = \int_{-\infty}^{x_k} f_k(y_k)dy_k.
\eeq 
For instance, for $k=n$,  using eq. \eqref{eq:distribution_xk_gaussian_final}:
\beq
F_n(x) = \int_{-\infty}^{x} \frac{n}{2^{n-1}\sigma\sqrt{2\pi}} \text{erfc}(-\frac{y}{\sigma\sqrt{2}})^{n-1}e^{-\frac{y^2}{2\sigma^2}} dy
\eeq
By using the substitution $u = -y$, $du = -dy$:
\beq
F_n(x) & = -\int_{+\infty}^{-x} \frac{n}{2^{n-1}\sigma\sqrt{2\pi}} \text{erfc}(\frac{u}{\sigma\sqrt{2}})^{n-1}e^{-\frac{u^2}{2\sigma^2}} du \\
& =\int_{-x}^{+\infty} \frac{n}{2^{n-1}\sigma\sqrt{2\pi}} \text{erfc}(\frac{u}{\sigma\sqrt{2}})^{n-1}e^{-\frac{u^2}{2\sigma^2}} du \\
& = \frac{1}{2^n} \text{erfc}(-\frac{x}{\sigma \sqrt{2}})^n
\eeq 
We can then make the approximation that for any $x \leq 0$, $F_n(x) << 1 - $.  For the first integral of \eqref{eq:expected_value_integral}, by using the same substitutions as in \autoref{sec:distribution_max_abs_gaussian}, we get
\beqq
&\int_{0}^{+\infty} 1 - \frac{1}{2^n} \text{erfc}(-\frac{x}{\sigma \sqrt{2}})^n dx = \int_{0}^{+\infty} 1 - \frac{1}{2^n} \left( 1-\text{erf}(-\frac{x}{\sigma \sqrt{2}}) \right)^n dx = \int_{0}^{+\infty} 1 - \frac{1}{2^n} \left( 1+\text{erf}(\frac{x}{\sigma \sqrt{2}}) \right)^n dx \\
&\approx  \frac{\sigma\sqrt{2\pi}}{2}\int_{0}^{+\infty} 1 - \frac{1}{2^n} \left( 1+ \sqrt{1-e^{-y^2}} \right)^n dy  = \frac{\sigma\sqrt{2\pi\ln(n)}}{2}  \int_{0}^{+\infty} 1 - \frac{1}{2^n} \left( 1+ \sqrt{1-\frac{1}{n^{v^2}}} \right)^n dv \\
&\approx \frac{\sigma\sqrt{2\pi\ln(n)}}{2} \int_{0}^{+\infty} 1 - \frac{1}{2^n} \left( 1+ 1-\frac{1}{2n^{v^2}} \right)^n dv \approx \frac{\sigma\sqrt{2\pi\ln(n)}}{2} \int_{0}^{+\infty} 1 - \left( 1-\frac{1}{4n^{v^2}} \right)^n dv
\eeqq
Using the convergence (see \autoref{sec:distribution_max_abs_gaussian}), we obtain the leading order term
\beq\label{eq:expected_value_xn_gaussian}
    \int_{0}^{+\infty} 1 - \frac{1}{2^n} \text{erfc}(-\frac{x}{\sigma \sqrt{2}})^n  \approx  \frac{\sigma\sqrt{2\pi\ln(n)}}{2}.
\eeq
The second integral of \eqref{eq:expected_value_integral} can be computed similarly:
\beq 
\int_{-\infty}^{0} \frac{1}{2^n} \text{erfc}(-\frac{x}{\sigma \sqrt{2}})^n dx & = \int_{-\infty}^{0} \frac{1}{2^n} \left( 1-\text{erf}(-\frac{x}{\sigma \sqrt{2}}) \right)^n dx \\
& \approx  \frac{\sigma\sqrt{2\pi}}{2}\int_{-\infty}^{0} \frac{1}{2^n} \left( 1 - \sqrt{1-e^{-y^2}} \right)^n dy \\ & = \frac{\sigma\sqrt{2\pi\ln(n)}}{2}  \int_{-\infty}^{0} \frac{1}{2^n} \left( 1 - \sqrt{1-\frac{1}{n^{v^2}}} \right)^n dv \\
& \approx \frac{\sigma\sqrt{2\pi\ln(n)}}{2} \int_{-\infty}^{0} \frac{1}{2^n} \left( 1 - 1-\frac{1}{2n^{v^2}} \right)^n dv 
 \\ & \approx \frac{\sigma\sqrt{2\pi}}{2} \int_{-\infty}^{0} \frac{\sqrt{\ln(n)}}{2^n}\left(-\frac{1}{4n^{v^2}} \right)^n dv
\eeq 

Since \[
  \frac{\sqrt{\ln(n)}}{2^n}\left(-\frac{1}{4n^{v^2}} \right)^n
  \;\longrightarrow\;
  0,
\]
dominated convergence yields
\[
  \int_{-\infty}^{0} \frac{1}{2^n} \text{erfc}(-\frac{x}{\sigma \sqrt{2}})^n dx
  \;\longrightarrow\; 0.
\]
We therefore obtain the expected value of the maximum among $n$ Gaussian variables:
\beq\label{eq:expected_value_xn_gaussianpp}
    \eE(X_{(n)}) \approx  \frac{\sigma\sqrt{2\pi\ln(n)}}{2}.
\eeq
For any $k$,we rewrite the density function $f_k$ as 
\beq
f_k(x) & = \frac{n!}{(k-1)!(n-k)!} \frac{\text{erfc}(-\frac{x}{\sigma\sqrt{2}})^{k-1} \text{erfc}(\frac{x}{\sigma\sqrt{2}})^{n-k}e^{-\frac{x^2}{2\sigma^2}}}{2^{n-1}\sigma\sqrt{2\pi}} \\
& = \frac{n!}{(k-1)!(n-k)!} F(x)^{k-1}(1-F(x)^{n-k} f(x),
\eeq 
where $\ds f(x)=\frac{e^{-\frac{x^2}{2\sigma^2}}}{\sigma\sqrt{2 \pi}}$ and $\ds F(x) =  \frac{\text{erfc}(-\frac{x}{\sigma\sqrt{2}})}{2}$ are respectively the density and cumulative distribution functions of a Gaussian law $\mathcal{N}(0, \sigma^2).$
We can rewrite $f_k$ as 
\beqq
f_k(x) & = \ds\frac{n!}{(k-1)!(n-k)!} F(x)^{k-1}(1-F(x)^{n-k} f(x) \\
& =  \frac{n!}{(k-1)!(n-k)!} f(x)F(x)^{k-1}\sum_{i=0}^{n-k}\binom{n-k}{i}(-1)^{n-k-i}F(x)^{n-k-i} \\
& = \frac{n!}{(k-1)!(n-k)!}\sum_{i=0}^{n-k}\binom{n-k}{i}(-1)^{n-k-i}F(x)^{n-i-1}f(x).
\eeqq
This is equivalent to  the cumulative form: 
\beqq
1 - F_k(x) = P(X_{(k)} \geq x) & = \int_{x}^{+\infty}\frac{n!}{(k-1)!(n-k)!}\sum_{i=0}^{n-k}\binom{n-k}{i}(-1)^{n-k-i}F(x)^{n-i-1}f(x) \\
& = \frac{n!}{(k-1)!(n-k)!} \sum_{i=0}^{n-k}\binom{n-k}{i}(-1)^{n-k-i} \int_{x}^{+\infty}F(x)^{n-i-1}f(x) \\
& = \frac{n!}{(k-1)!(n-k)!} \sum_{i=0}^{n-k}\binom{n-k}{i}(-1)^{n-k-i} \frac{1-F(x)^{n-i}}{n-i}. 
\eeqq
Finally, 
\beq
F_k(x) & = \int_{-\infty}^{+x}\frac{n!}{(k-1)!(n-k)!}\sum_{i=0}^{n-k}\binom{n-k}{i}(-1)^{n-k-i}F(x)^{n-i-1}f(x) \\
& = \frac{n!}{(k-1)!(n-k)!} \sum_{i=0}^{n-k}\binom{n-k}{i}(-1)^{n-k-i} \frac{F(x)^{n-i}}{n-i}.
\eeq 
We conclude that for $n-k\ll  n$ and  $i\ll n$, we can use similar arguments than for $k=n$ to prove that $\ds \int_{-\infty}^{0} F(x)^{n-i}\longrightarrow\ 0$. Hence,
\beq
E(X_{(k)}) & = \int_0^{+\infty} (1-F_k(x))dx - \int_{-\infty}^0F_k(x)dx  \approx \int_0^{+\infty} (1-F_k(x))dx \\
& = \int_0^{+\infty}  \frac{n!}{(k-1)!(n-k)!} \sum_{i=0}^{n-k}\binom{n-k}{i}(-1)^{n-k-i} \frac{1-F(x)^{n-i}}{n-i} dx \\
& =\frac{n!}{(k-1)!(n-k)!} \sum_{i=0}^{n-k}\binom{n-k}{i}(-1)^{n-k-i}\frac{1}{n-i} \int_0^{+\infty}1-F(x)^{n-i}dx\\
& \approx \frac{n!}{(k-1)!(n-k)!} \sum_{i=0}^{n-k}\binom{n-k}{i}(-1)^{n-k-i}\frac{1}{n-i}\frac{\sigma\sqrt{2\pi\ln(n-i)}}{2}.
\eeq 
Finally, we can conclude that the expected value of $X_{(k)}$ when $n-k << n$ can be approximated by 
\beq\label{eq:expected_value_xk_gaussian}
    \eE(X_{(k)}) \approx  \frac{\sigma\sqrt{2\pi}n (n-1) \dots (n-(n-k))}{2}\sum_{i=0}^{n-k}(-1)^{n-k-i}\frac{1}{i!(n-k-i)!(n-i)}\sqrt{\ln(n-i)}.
\eeq
%%%%%%%%%%%%%%%%%%%%%%%%%%%%%%%%%%%%%%%%%%%%%%%%%%%%%%%
\section{Acknowledgements}
We thank S. Majumdar for discussions on this manuscript and pointing out his recent book as a reference. 
%%%%%%%%%%%%%%%%%%%%%%%%%%%%%%%%%%%%
\bibliographystyle{ieeetr}
%%%%%%%%%%%%%%%%%%%%%%%%%%%%%%%%%%%%%%%%%%%%%%%%%%%%%%%%%%%%%%%%%%%%%%%
\bibliography{ref.bib}
\end{document}